\documentclass[twoside]{article}

\usepackage[accepted]{aistats2025}
\usepackage[round]{natbib}

\usepackage{longtable}
\usepackage{appendix}
\usepackage{hyperref}
\usepackage{algorithm}
\usepackage{algcompatible}
\usepackage{algpseudocode}
\usepackage{multirow}
\usepackage{amsmath}
\usepackage{bm}
\usepackage{amssymb}
\usepackage{graphicx}
\usepackage{subcaption}
\usepackage{dsfont}
\usepackage{wrapfig}
\usepackage{xcolor}
\usepackage{amsthm}
\usepackage{enumitem}

\newtheorem{proposition}{Proposition}[section]

\newtheorem{lemma}{Lemma}[section]
\newtheorem{definition}{Definition}[section]

\begin{document}
%

%
\runningauthor{Shuhui Zhu, Baoxiang Wang, Sriram Ganapathi Subramanian, Pascal Poupart}

\twocolumn[

\aistatstitle{Learning to Negotiate via Voluntary Commitment}
\aistatsauthor{Shuhui Zhu \And Baoxiang Wang}
\aistatsaddress{University of Waterloo \\ Vector Institute 
\\ \texttt{shuhui.zhu@uwaterloo.ca} 
\And  The Chinese University of Hong Kong, Shenzhen 
\\ \texttt{bxiangwang@cuhk.edu.cn}
}
\aistatsauthor{Sriram Ganapathi Subramanian \And Pascal Poupart}
\aistatsaddress{Vector Institute 
\\ \texttt{sriram.subramanian@vectorinstitute.ai}
\And University of Waterloo \\ Vector Institute 
\\ \texttt{ppoupart@uwaterloo.ca}
}

]


\begin{abstract}
The partial alignment and conflict of autonomous agents lead to mixed-motive scenarios in many real-world applications. However, agents may fail to cooperate in practice even when cooperation yields a better outcome. One well known reason for this failure comes from non-credible commitments. To facilitate commitments among agents for better cooperation, we define Markov Commitment Games (MCGs), a variant of commitment games, where agents can voluntarily commit to their proposed future plans. Based on MCGs, we propose a learnable commitment protocol via policy gradients. We further propose incentive-compatible learning to accelerate convergence to equilibria with better social welfare. Experimental results in challenging mixed-motive tasks demonstrate faster empirical convergence and higher returns for our method compared with its counterparts. Our code is available at \url{https://github.com/shuhui-zhu/DCL}.
\end{abstract}

\section{Introduction}
In mixed-motive applications~\citep{dafoe2020open}, agents often fail to cooperate even when cooperation leads to better outcomes. One key reason is the issue of non-credible commitments. For instance, in the Prisoner's Dilemma (Table~\ref{PD_matrix}), mutual cooperation would lead to higher payoffs for both players compared to mutual defection, but each player, driven by its self-interest, is incentivized to defect regardless of the other's choice. As a result, credible commitments to cooperate cannot be established. 

\begin{figure*}[t]
    \centering
    \includegraphics[width=0.85\textwidth]{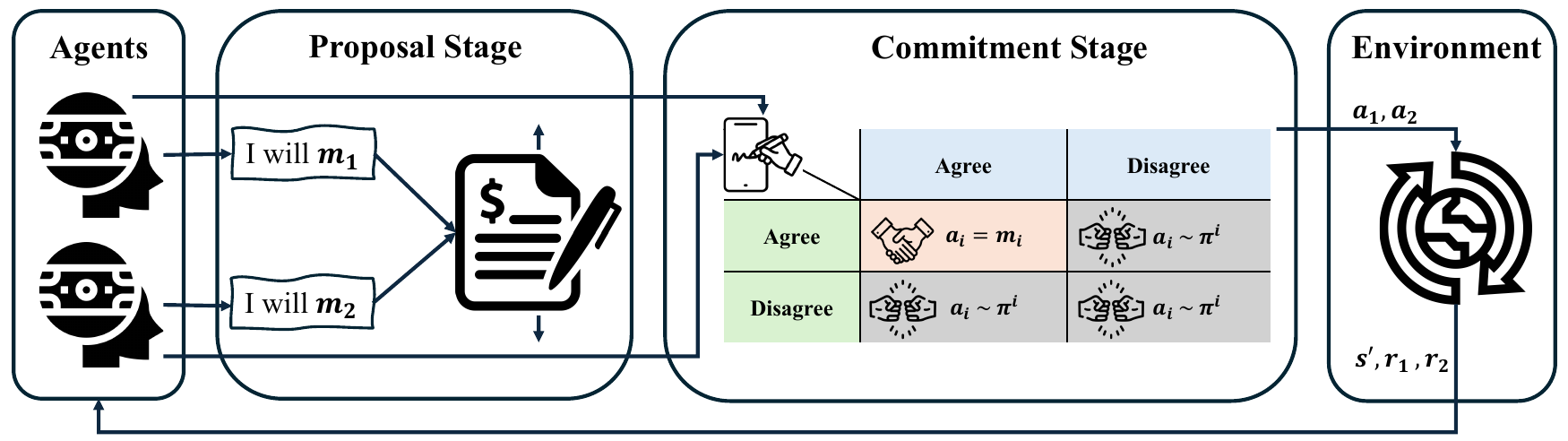}
    \caption{Markov Commitment Game: A Markov commitment game consists of three stages. In the first stage, agents announce their proposed future actions. In the second stage, agents observe others' proposals and decide whether to commit to the joint plan. In the final stage, agents choose their actions: if all agents commit, they follow their proposals; if any agent does not commit, all agents independently select actions based on the current state. Afterward, agents observe the resulting rewards and transit to the next state.}
    \label{MSMG}
    \vspace*{-0.2cm}
\end{figure*}
To mitigate the commitment problem, a commitment device~\citep{165211,sun2023cooperative} is often required to ensure that agents fulfill their commitments, either by binding their actions to fixed strategies~\citep{schelling1980strategy,renou2009commitment,kalai2010commitment,digiovanni2023commitment} or imposing penalties for noncompliance~\citep{bryan2010commitment}. 
In particular, conditional commitment devices~\citep{kalai2010commitment,dafoe2020open} have been verified to enhance cooperation in the Prisoner's Dilemma. 
When one player conditionally commits to cooperate if and only if the other does the same, the other player is motivated to cooperate.
However, these conditional commitment mechanisms, tailored to specific problems, typically rely on fixed, pre-specified rules, leaving no room for adaptation in more complex, dynamic environments. 
Additionally, such mechanisms are designed primarily for simple, repeated games such as the Prisoner's Dilemma, limiting their applicability to a broader range of strategic scenarios where the conditions for cooperation may evolve over time. 

To address these limitations,  we propose a learnable commitment mechanism, named differentiable commitment learning (DCL) based on the introduced Markov Commitment Games (MCGs, Figure~\ref{MCG}). MCGs are a variant of commitment games~\citep{renou2009commitment,bryan2010commitment,forges2013folk,digiovanni2023commitment}. In two-phase commitment games, each agent first announces a unilateral commitment to a subset of possible strategies, then selects an action based on strategies they have committed to. 
Different from commitment games, MCGs incorporate an additional proposal phase, where agents release a proposed future plan of their own actions in the current state without disclosing their strategies for other states. 
As a result, MCGs do not require mutual transparency of commitment strategies and avoid incompatibilities in commitment implementation. Furthermore, commitments in MCGs have linear size in the planning horizon and are therefore more tractable for agents to reason through, whereas in conditional commitment games~\citep{bryan2010commitment,forges2013folk,digiovanni2023commitment}, commitments are recursive and potentially infinite.

The core idea of DCL in MCGs is to learn a commitment protocol that enables agents to voluntarily align their actions based on the commitments of others. Under the assumption of self-interested agents, DCL adopts the scheme of reinforcement learning~\citep{sutton2018reinforcement}, optimizes long-term individual returns via policy gradients. Different from common RL algorithms that treat other agents as part of the environment, DCL allows backpropagation through actual or estimated policies of other agents.
The advantages of DCL are twofold. 
1) The commitment mechanism is agnostic to environment dynamics so that it can generalize across various tasks. Whereas in commitment games~\citep{renou2009commitment,bryan2010commitment,forges2013folk,digiovanni2023commitment}, the commitment strategies are pre-defined for specific problems. 2) DCL provides more accurate value evaluation and policy gradient estimations through backpropagation across commitment channels. By explicitly leveraging the interdependence of agents' decisions, DCL enhances learning outcomes. Whereas other baseline RL algorithms~\citep{schulman2017proximal,haupt2022formal,ivanov2023mediated} treat other agents as part of the environment, resulting in non-stationarity from each agent's perspective.

Extensive experiments in tabular, sequential and iterative social dilemmas verify the efficiency of our approach in promoting cooperation. DCL significantly outperforms several baseline methods, including independent RL, contract-based reward transfer RL, and mediated multi-agent RL, often by establishing mutually beneficial multilateral commitments.
\section{Related Works}
\vspace*{-0.1cm}
\subsection{Binding Contracts Mechanism}
Binding contracts are generally applied to establish commitments in multi-agent systems. 
The literature offers various approaches to contract design.
\citet{wang2024deep,han2017evolution,sandholm1996advantages} developed contracts that bind agents' future actions through side payments, rewarding agents for fulfilling commitments and penalizing them for noncompliance. \citet{haupt2022formal,sodomka2013coco} also explored mechanisms where agents voluntarily agree to binding reward transfers. However, these methods directly alter agents' incentives, which may not be feasible in practice.

Instead, \citet{kramar2022negotiation,de2020strategic,hughes2020learning} proposed adaptive binding actions without reward transfers, which are similar to MCGs but differ in specific details. \citet{de2020strategic} focused on turn-taking games with unilateral commitments, while MCGs emphasize simultaneous moves and multilateral commitments. \citet{hughes2020learning} required agents to propose a joint plan for all, with multilateral commitment only if they propose the same plan. In MCGs, however, each agent proposes an individual plan and uses a separate commitment model to decide whether to commit or not. \citet{kramar2022negotiation} introduced pairwise negotiation through Nash Bargaining Solution (NBS)~\citep{binmore1986nash}, aiming to maximize the product of agents' utilities. In contrast, MCGs focus on selfish agents aiming to maximize their individual long-term returns.



\subsection{Altruistic Third Party}
Without manipulating agents' rewards, \citep{ivanov2023mediated,mcaleer2021improving, greenwald2003correlated} introduced pro-social third parties to mediate agents' actions and induce cooperative behaviors. These approaches optimize social welfare such as the sum of agents' returns while incorporating rationality constraints that define equilibria, ensuring that self-interested agents have no incentive to deviate from their strategies. 
Specifically, utilitarian correlated-Q learning~\citep{greenwald2003correlated} utilized a centralized model to optimize the joint action probability distribution of all agents, with an objective that maximizes the sum of the agents’ rewards.
In contrast, \citet{ivanov2023mediated,mcaleer2021improving} trained agents to optimize their individual payoffs, allowing them to follow the recommendations of a prosocial mediator or take their actions independently if those recommendations do not align with their self-interests. However, these approaches still rely on a centralized altruistic third party, which may become ineffective in highly conflicting environments where collective interests significantly clash with individual self-interests. 
\section{Background on Commitment Games}
A normal form game $G = (\mathcal{N}, (\mathcal{R}^i , \mathcal{A}^i)_{i \in \mathcal{N}})$ consists of a set $\mathcal{N}$ of agents, where each agent $i$ chooses an action $a^i \in \mathcal{A}^i$ and earns a reward according to the function $\mathcal{R}^i : \prod_j \mathcal{A}^j \rightarrow \mathbb{R}$.  A commitment game \citep{renou2009commitment} extends a normal form game to two phases where each agent first makes a commitment and then plays an action.  Formally, a commitment game $CG = (\mathcal{N}, (\mathcal{R}^i , \mathcal{A}^i, \mathcal{C}^i)_{i \in N})$ extends a normal form game with a commitment space $\mathcal{C}^i$ for each agent.  Player $i$'s strategy $(c^i,\sigma^i)$ consists of a commitment $c^i\in\mathcal{C}^i$ and a response function $\sigma^i : \prod_j \mathcal{C}^j \rightarrow \mathcal{A}^i$.  For example, \citet{renou2009commitment} considered unconditional unilateral commitments where a commitment $c^i \subseteq \mathcal{A}^i$ is a subset of the action space, meaning that the agent commits to choose an action in that subset.  Such unconditional unilateral commitments can yield better equilibria (i.e., Pareto optimal) when ruling out some threats will incite other agents to cooperate.  However, in other games such as Prisoner's Dilemma, no unilateral commitment will induce convergence to mutual cooperation.  

\citet{kalai2010commitment} proposed conditional unilateral commitments $\mathcal{C}^i : \prod_{j\neq i} \mathcal{C}^j \rightarrow \mathcal{A}^i$, where agents commit to some actions conditioned on the commitments of others.  This space of commitments is recursive and potentially infinite, however it can turn mutual cooperation into a stable equilibrium in Prisoner's Dilemma when both agents commit to cooperating conditioned on the other one cooperating too.
\citet{kalai2010commitment} further augmented conditional unilateral commitments with a voluntary commitment space. In this voluntary commitment space, agents are allowed to play the normal form game $G$ without making any advanced commitment. Thus, agents will independently select their actions $a^i \in A^i$ if they voluntarily decide not to commit to any $c^i \in \mathcal{C}^i$. However, this conditional commitment mechanism requires agents to reveal their commitment strategies~\citep{kalai2010commitment,forges2013folk} or source code of their models~\citep{digiovanni2023commitment}, which may be impractical and lead to incompatibilities in commitment implementation.  Two tables in the supplementary material 
summarize the differences and similarities between various types of games and associated algorithms to optimize strategies.

\section{Markov Commitment Games}
The ability to make binding commitments is a fundamental mechanism for promoting cooperation. To enable strategic commitment-making among intelligent agents in multi-agent systems, we formulate a Markov Commitment Game (MCG, Figure~\ref{MSMG}), formally defined by a tuple 
\begin{equation}
\label{MCG}
    MCG = (\mathcal{N}, \mathcal{S},  \mathcal{T}, (\mathcal{M}^i,\mathcal{C}^i,\mathcal{A}^i,\mathcal{R}^i)_{i\in \mathcal{N}},\gamma).
\end{equation}
MCGs include three stages. At each time step $t$, the agent $i \in \mathcal{N}$ observes a global state $s_t \in \mathcal{S}$ and announces a proposal $m^i \in \mathcal{M}^i=\mathcal{A}^i$ in the first stage. Then each agent $i$ observes the joint proposal $\mathbf{m}=(m^i)_{i\in \mathcal{N}}$ and makes a commitment decision $c^i \in \mathcal{C}^i=\{0,1\}$ in the second stage, where $c^i=1$ indicates that agent $i$ commits to the joint proposal, $c^i=0$ indicates that agent $i$ rejects the joint proposal. In the third stage, if all agents commit to the joint plan, they execute the actions in the proposal, i.e., $a^i=m^i, \forall i\in \mathcal{N}$; otherwise, each agent $i$ independently selects an action $a^i \in \mathcal{A}^i$. Agent $i$ receives the reward $r^i$, determined by the reward function $\mathcal{R}^i: \mathcal{S} \times \mathcal{A} \rightarrow \mathbb{R}$, where $\mathcal{A}=(\mathcal{A}^i)_{i\in \mathcal{N}}$ represents the joint action space. Meanwhile, the next state $s_{t+1}$ is generated by the transition function $\mathcal{T}: \mathcal{S} \times \mathcal{A} \rightarrow \Delta(\mathcal{S})$, which satisfies the Markov property and the stationarity condition, i.e., $\mathcal{T}(s_{t+1}=s'|s_t=s, \mathbf{a}_t=\mathbf{a})=\mathcal{T}(s_{t+1}=s'|s_t=s,\mathbf{a}_t=\mathbf{a},s_{t-1},\mathbf{a}_{t-1}...,s_0, \mathbf{a}_0)=\mathcal{T}(s'|s,\mathbf{a}), \forall t$. This process is repeated until the episode ends. It is important to note that the transition distribution conditions on the current state and joint actions only, not on the proposals or commitment decisions. This is because proposals and commitments indirectly influence the transition by affecting the actions executed. 

In an MCG, each agent has three decisions to make at each time step: what to propose, whether to commit or not, and how to choose actions without joint commitment. Therefore, we decompose each agent's behavioral model into three strategic policies. 
The proposal policy, $\phi^i_{\eta^i}: \mathcal{S} \rightarrow \Delta(\mathcal{M}^i)$, maps the current state $s_t$ to a distribution over agent $i$' space of proposals. The commitment policy, $\psi^i_{\zeta^i}: \mathcal{S} \times \mathcal{M} \rightarrow \Delta(\mathcal{C}^i)$, depends on the state $s_t$ and the joint proposal $\mathbf{m}_t \in \mathcal{M}=(\mathcal{M}^i)_{i\in \mathcal{N}}$. The action policy, $\pi^i_{\theta^i}: \mathcal{S} \rightarrow \Delta(\mathcal{A}^i)$, samples action based on the current state $s_t$ only.

MCGs adopt a strategic commitment mechanism in mixed-motive multi-agent systems. 
In this framework, the environment also serves as a commitment device, enforcing agents' voluntarily imposed restrictions on their future actions. 
Agents in MCGs have access to this device, which is effective only when all self-interested agents agree to commit to a public joint plan. If any agent declines, all agents will independently select actions without restrictions by commitment. Thus, the commitment device facilitates a conditional commitment: agents agree to execute their proposed actions only if every other agent also commits to the joint plan.

Driven by self-interest, the objective of each agent $i$ is to find the optimal strategy $(\phi^{i}_{\eta^{i*}}, \psi^{i}_{\zeta^{i*}}, \pi^{i}_{\theta^{i*}})$ that maximizes their future expected return, i.e. the expected cumulative discounted reward, defined by
\begin{equation}
\label{selfish_objective}
    \max_{\eta^i, \zeta^i, \theta^i} V^i_{\bm{\phi,\psi,\pi}}(s)=\mathbb{E}_{\bm{\phi,\psi,\pi}}[\sum_{k=t}^\infty \gamma^{k-t} r_{k+1}^i|s_t=s],
\end{equation}
where $\gamma$ is the discounted factor, $\bm{\phi}=(\phi^i_{\eta^i})_{i\in \mathcal{N}}$, $\bm{\psi}=(\psi^i_{\zeta^i})_{i\in \mathcal{N}}$, $\bm{\pi}=(\pi^i_{\theta^i})_{i\in \mathcal{N}}$. Note that agent $i$'s value function $V^i_{\bm{\phi,\psi,\pi}}(s)$ is dependent on other agents' strategies, as the collective actions of all agents jointly decide the rewards and state transitions in multi-agent systems. Meanwhile, each agent's proposal and commitment decision also indirectly affect others' expectation of their future returns. 
Therefore, the impact of other players' policies on each agent's objective should be properly evaluated during learning.
\subsection{Equilibrium Analysis in Prisoner's Dilemma}
MCGs induce a conditional commitment mechanism, which can lead to different strategic behaviors and outcomes compared to a game without such commitments. 
\begin{proposition}
\label{equilibrium_pd}
    Mutual cooperation is a Pareto-dominant Nash equilibrium in the MCG of the Prisoner's Dilemma. 
\end{proposition}
Specifically, we demonstrate with Proposition~\ref{equilibrium_pd} that with the ability to commit, both players have an incentive to strategically propose and commit to cooperation, given the other agent does the same, thereby transforming mutual cooperation into a Pareto-dominant Nash equilibrium.
The formal proof of this proposition is provided in Appendix~\ref{pd_proof}. 
\section{Differentiable Commitment Learning}
Based on MCGs, we propose differentiable commitment learning (DCL) under the assumption of self-interested agents. Instead of treating other agents as part of the environment, DCL considers joint actions when evaluating individual returns. To formulate this idea, we define the state-action value function of agent $i$ in MCGs as $Q^i_{\bm{\phi,\psi,\pi}}(s,\mathbf{a})=\mathbb{E}_{\bm{\phi,\psi,\pi}}[\sum_{k=t}^\infty \gamma^{k-t} r_{k+1}^i|s_t=s,\mathbf{a}_t=\mathbf{a}]$, representing the expected future returns conditioned on the current state and the joint actions. 
Because the environment's transitions and reward function in MCGs depend only on the state and joint actions, the state-action value function does not condition on proposals or commitments either. Under the scheme of on-policy reinforcement learning~\citep{sutton2018reinforcement}, DCL estimates this state-action value function by minimizing the mean square error between 
$Q^i_{\bm{\phi,\psi,\pi}}(s, \mathbf{a})$
and the Monte Carlo returns $\hat{G}_t^i=\sum_{k=t}^T\gamma^{k-t}r_{k+1}^i$ of the sampled trajectories.
Similar to the policy gradient theorem~\citep{sutton1999policy}, we then derive unbiased policy gradients based on $Q^i_{\bm{\phi,\psi,\pi}}(s, \mathbf{a})$ in Equations~(\ref{action_gradient}),~(\ref{commitment_gradient}), and (\ref{proposal_gradient}) respectively. The complete proof of Lemma~\ref{gradient_derivation_lemma} is provided in Appendix~\ref{gradient_derive}.
\begin{lemma}
\label{gradient_derivation_lemma}
Given proposal policy $\phi^i_{\eta^i}$, commitment policy $\psi^i_{\zeta^i}$ and the action policy $\pi^i_{\theta^i}$ of each agent $i$ in an MCG~(\ref{MCG}), the gradients of the value function $V^i_{\bm{\phi,\psi,\pi}}(s)$ w.r.t. $\theta^i$, $\zeta^i$, $\eta^i$ are 
{\small
\begin{equation}
\label{action_gradient}
\begin{aligned}
    \nabla_{\theta^i} V^i_{\bm{\phi,\psi,\pi}}(s) \propto &\mathbb{E}_{x \sim \rho_{\bm{\phi,\psi,\pi}}, \mathbf{m} \sim \bm{\phi}, \mathbf{c} \sim \bm{\psi}, \mathbf{a}\sim\bm{\pi}}\Big[\Big(1-\mathds{1}(\mathbf{c=1})\Big)\\
    &Q_{\bm{\phi,\psi,\pi}}^i(x,\mathbf{a})\nabla_{\theta^i}\log \pi^i(a^i|x) \Big],
\end{aligned}
\end{equation}}
{\small
\begin{equation}
\label{commitment_gradient}
\begin{aligned}
    &\nabla_{\zeta^i} V^i_{\bm{\phi,\psi,\pi}}(s)\\
    \propto &\mathbb{E}_{x \sim \rho_{\bm{\phi,\psi,\pi}}, \mathbf{m} \sim \bm{\phi}, \mathbf{c} \sim \bm{\psi}, \mathbf{a}\sim\bm{\pi}}\Bigg[\Big[\mathds{1}(\mathbf{c=1})Q_{\bm{\phi,\psi,\pi}}^i(x,\mathbf{m})\\
    &+\Big(1-\mathds{1}(\mathbf{c=1})\Big)Q_{\bm{\phi,\psi,\pi}}^i(x,\mathbf{a})\Big]\nabla_{\zeta^i}\log\psi^i(c^i|x, \mathbf{m})\\
    &+ \Big[Q_{\bm{\phi,\psi,\pi}}^i(x,\mathbf{m})-Q_{\bm{\phi,\psi,\pi}}^i(x,\mathbf{a}) \Big]\prod_{k \neq i}\mathds{1}(c^{k}=1)\\
    &\cdot \nabla_{\zeta^i}\mathds{1}(c^i=1)
    \Bigg],
\end{aligned}
\end{equation}}
{\small
\begin{equation}
\label{proposal_gradient}
\begin{aligned}
    &\nabla_{\eta^i} V^i_{\bm{\phi,\psi,\pi}}(s)\\
    \propto &\mathbb{E}_{x \sim \rho_{\bm{\phi,\psi,\pi}}, \mathbf{m} \sim \bm{\phi}, \mathbf{c} \sim \bm{\psi}, \mathbf{a}\sim\bm{\pi}}\Bigg[
        \Big[\mathds{1}(\mathbf{c=1})Q_{\bm{\phi,\psi,\pi}}^i(x,\mathbf{m})\\
        &+\Big(1-\mathds{1}(\mathbf{c=1})\Big)Q_{\bm{\phi,\psi,\pi}}^i(x,\mathbf{a}) \Big]\\
        &\cdot\Big(\nabla_{\eta^i}\log\phi^i(m^i|x)+
        \sum_j\nabla_{\eta^i}\log\psi^j(c^j|x, \mathbf{m}) \Big)\\
        &+\sum_{j}\prod_{k\neq j} \mathds{1}(c^k=1)\Big[Q_{\bm{\phi,\psi,\pi}}^i(x,\mathbf{m})-Q_{\bm{\phi,\psi,\pi}}^i(x,\mathbf{a}) \Big]\\
        &\cdot\nabla_{\eta^i}\mathds{1}(c^j=1)
        \Bigg],
\end{aligned}
\end{equation}}
where $\mathds{1}(\cdot)$ denotes the indicator function, which equals $1$ if the condition inside is true and $0$ otherwise; $\rho_{\bm{\phi,\psi,\pi}}(x)$ denotes a discounted probability of state $x$ encountered, starting at $s$ and then with all agents following $\bm{\phi,\psi,\pi}: \rho_{\bm{\phi,\psi,\pi}}(x) = \sum_{t=0}^{\infty} \gamma^t Pr\{s_t=x|s_0=s\}$. 
\end{lemma}
Through policy gradients in Lemma~\ref{gradient_derivation_lemma}, DCL enables agents to optimize their strategies by considering both direct and indirect effects of their policies on their utilities. 
To capture the direct impact, DCL allows agents to differentiate through their own policies, updating in the direction that maximizes their individual returns. On the other hand, DCL allows agents to consider how their decisions influence others' commitments and how these influences, in turn, affect their own utilities. This indirect influence is leveraged by differentiation through the commitment policies of other players when computing $\nabla_{\eta^i} V^i_{\bm{\phi,\psi,\pi}}(s)$. 
To backpropagate through discrete commitments, we apply the Gumbel-Softmax distribution~\citep{jang2016categorical} for differentiable sampling.

Instead of limiting DCL to centralized training (Appendix~\ref{Centralized_DCL_details}) with access to other agents' policies, we extend DCL to fully decentralized settings (Appendix~\ref{Decentralized_DCL_details}). In decentralized DCL, each agent estimates others' policies and differentiates through these estimates to update their own policies.
\begin{algorithm}[ht]
\caption{Differentiable Commitment Learning}\label{alg:dcl_summary}
\begin{algorithmic}
\State Input: initial parameters of action policy $\theta^i$, commitment policy $\zeta^i$, proposal policy $\eta^i$, action-value function $w^i$ for $i\in \mathcal{N}$, learning rate $\beta$, Lagrange multiplier $\lambda$, number of iterations $T$.
\For{k=$0,1,2, ..., T-1$}
    \State Collect set of trajectories $\mathcal{D}_k=\{\tau_t\}$ by running latest policies $(\theta^i,\zeta^i,\eta^i)$, $\forall i \in \mathcal{N}$.
    \State Compute Monte-Carlo discounted accumulative rewards $\hat{G}_t^i, \forall i\in\mathcal{N}$.
    \State Fit value function with gradient descent by minimizing the mean-squared error:
    \State {\tiny$$w^i_{k+1}=\arg \min_{w^i} \frac{1}{|\mathcal{D}_k|T}\sum_{\tau\in\mathcal{D}_k}\sum_{t=0}^T(Q_{w^i}^i(s_t,\mathbf{a}_t)-\hat{G}_t^i)^2.$$}
    \State Estimate action policy gradient $\hat{g}_{\theta^i_k}$ according to Equation~(\ref{action_gradient}).
    \State Estimate commitment policy gradient $\hat{g}_{\zeta^i_k}$ according to Equation~(\ref{commitment_gradient}).
    \State Estimate proposal policy gradient $\hat{g}_{\eta^i_k}$ w.r.t. expected return according to Equation~(\ref{proposal_gradient}).
    \State Estimate proposal policy gradient $\hat{g}_{\eta^i_k}^{'}$ w.r.t. the incentive-compatible constraints by
    {\tiny
    \State $$\frac{1}{|\mathcal{D}_k|}\sum_{\tau\in\mathcal{D}_k}\sum_{t=0}^T\sum_j \nabla_{\eta^i_k}\min \{0, Q^j_{w^j_{k+1}}(s_t, \mathbf{m}_t) - Q^j_{w^j_{k+1}}(s_t, \mathbf{a}_t) \}.$$}
    \State Update policy parameters for all agents with gradient ascent,
    \State {\tiny$$\theta^i_{k+1}=\theta^i_{k}+\beta \hat{g}_{\theta^i_k},\zeta^i_{k+1}=\zeta^i_{k}+\beta \hat{g}_{\zeta^i_k}, \eta^i_{k+1}=\eta^i_{k}+\beta \hat{g}_{\eta^i_k} + \lambda\hat{g}_{\eta^i_k}^{'}.$$}
\EndFor
\end{algorithmic}
\end{algorithm}
\subsection{Incentive-Compatible Constraints}
Although mutual cooperation can be a Nash equilibrium in MCGs for some mixed-motive environments, agents may still have the equilibrium selection problem when multiple equilibria exist. For instance, mutual defection is another Nash equilibrium of the MCG in Prisoner's Dilemma, with less pay-offs of both agents compared to mutual cooperation equilibrium in Lemma~\ref{gradient_derivation_lemma}. 
Even if agents are motivated by self-interest to select mutual cooperation equilibria over mutual defection equilibria with DCL, they may fail to find the equilibria with better outcomes because of inefficient exploration. 
To address this challenge, we introduce a set of incentive-compatible constraints on agents' proposal policies in Equation~(\ref{commitment_eq}), which encourage agents to find mutually beneficial proposals. 
\begin{equation}
\label{commitment_eq}
    \mathbb{E}_{\mathbf{m}\sim \bm{\phi}}[Q_{\bm{\phi,\psi,\pi}}^i(s, \mathbf{m})] \geq \mathbb{E}_{\mathbf{a}\sim \bm{\pi}}[Q_{\bm{\phi,\psi,\pi}}^i(s, \mathbf{a})]~~~\forall~i.
\end{equation}
Combining these incentive-compatible constraints with the self-interested objective, agents are driven to maximize their expected returns and propose mutually beneficial agreements. If a joint proposal results in outcomes worse than actions induced by independent action policy for any player, agents are penalized during training through a regularization term induced by constraints in Equation~(\ref{commitment_eq}). This regularization encourages agents to develop better agreements that benefit all players.
Meanwhile, these constraints do not sacrifice agents' self-interests, as they retain the ability to reject proposals that do not enhance their own utility.
Thus, they will follow their unconstrained policies unless a mutually beneficial agreement emerges. 

It is important to note that feasible solutions always exist for Equation~(\ref{commitment_eq}), as agents can align their proposal policies with their action policies, i.e. $\phi^i(s)=\pi^i(s)$ for $\forall i\in \mathcal{N}$. 
We then integrate these constraints into the objective function of agent $i$ with a Lagrange multiplier $\lambda$, to update the parameter $\eta^i$ of the proposal policy:
\begin{equation}
\begin{aligned}
\label{update_constrained_proposal}
    \eta^i \leftarrow &\eta^i + \nabla_{\eta^i} V^i_{\bm{\phi,\psi,\pi}}(s)+ \lambda \nabla_{\eta^i}\sum_j\min \{0, \\
    &\mathbb{E}_{\mathbf{m}\sim \bm{\phi}}[Q_{\bm{\phi,\psi,\pi}}^j(s, \mathbf{m})] - \mathbb{E}_{\mathbf{a}\sim \bm{\pi}}[Q_{\bm{\phi,\psi,\pi}}^j(s, \mathbf{a})] \}.
\end{aligned}
\end{equation}
Note that when $\lambda=0$, the proposal policies are not constrained by Equation~(\ref{commitment_eq}). 
The abstract pseudocode of DCL is provided in Algorithm~\ref{alg:dcl_summary}.
Please refer to Appendix~\ref{DCL_details} for more details about DCL.

\section{Experiments}
\begin{figure*}
\centering
\begin{subfigure}{1\textwidth}
  \centering
  \includegraphics[width=1\textwidth]{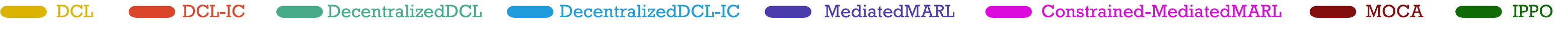}
\end{subfigure}%
\vspace{0.1mm}
\begin{subfigure}{.32\textwidth}
  \centering
  \includegraphics[width=1\textwidth]{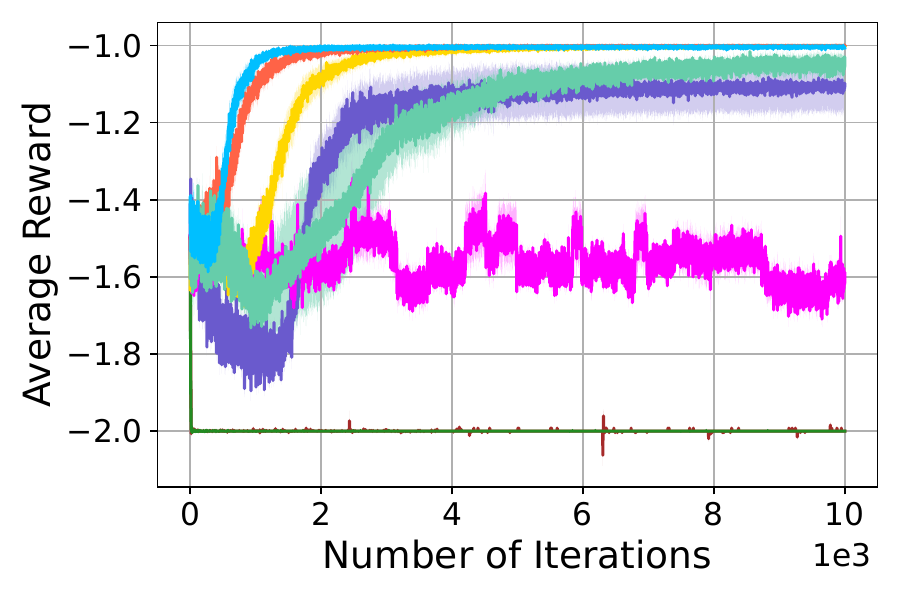}
  \caption{Agent $1$ Reward in PD}
  \label{fig:pd-a1}
\end{subfigure}%
\begin{subfigure}{.32\textwidth}
  \centering
  \includegraphics[width=1\textwidth]{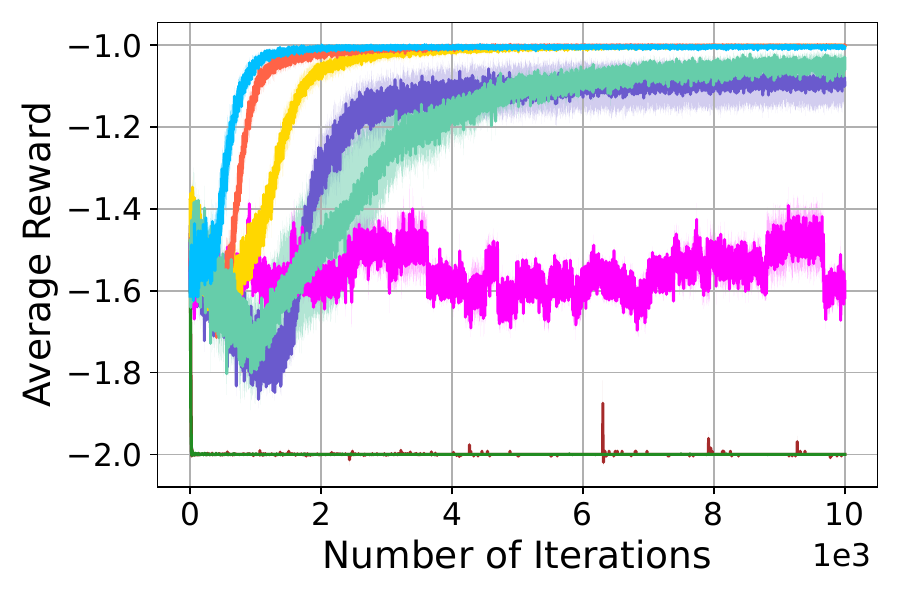}
  \caption{Agent $2$ Reward in PD}
  \label{fig:pd-a2}
\end{subfigure}%
\begin{subfigure}{.32\textwidth}
  \centering
  \includegraphics[width=1\textwidth]{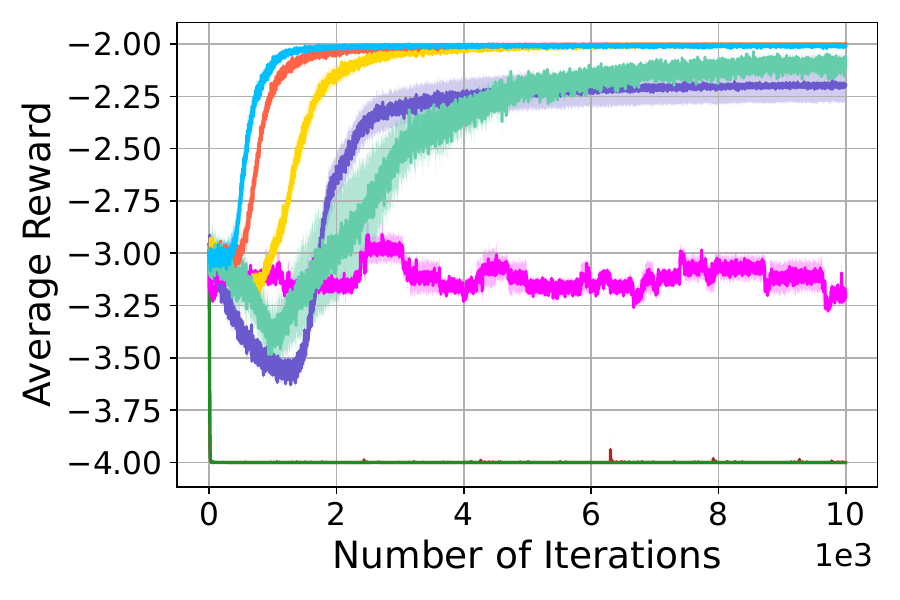}
  \caption{Social Welfare in PD}
  \label{fig:pd-social}
\end{subfigure}%
\caption{Prisoner's Dilemma: DCL v.s. Other Baselines}
\label{fig:pd-results}
\end{figure*}
\begin{figure*}
\centering
\begin{subfigure}{.32\textwidth}
  \centering
  \includegraphics[width=.98\textwidth]{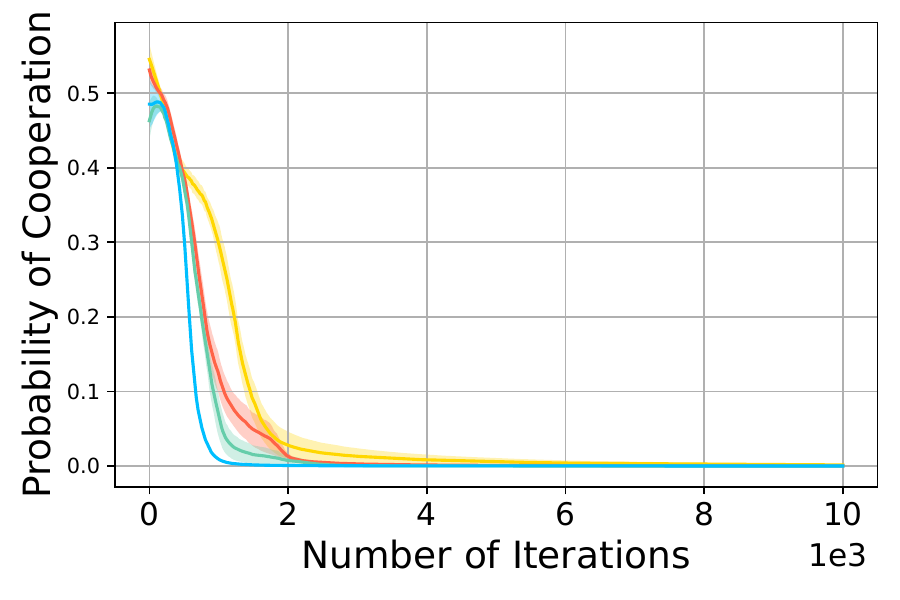}
  \caption{DCL Action Policy}
  \label{fig:pd-pi_a}
\end{subfigure}%
\begin{subfigure}{.32\textwidth}
  \centering
  \includegraphics[width=.98\textwidth]{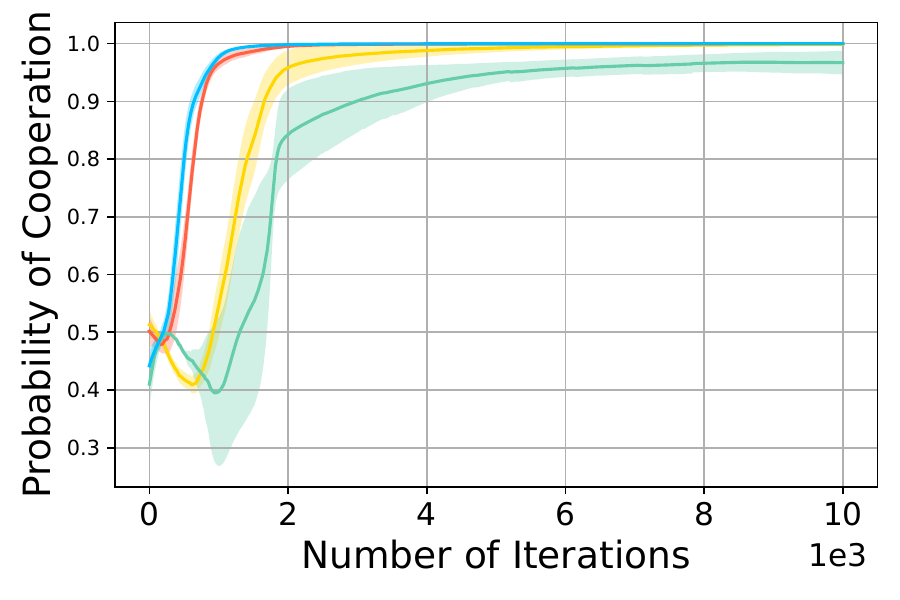}
  \caption{DCL Proposal Policy}
  \label{fig:pd-pi_m}
\end{subfigure}%
\begin{subfigure}{.32\textwidth}
  \centering
  \includegraphics[width=.98\textwidth]{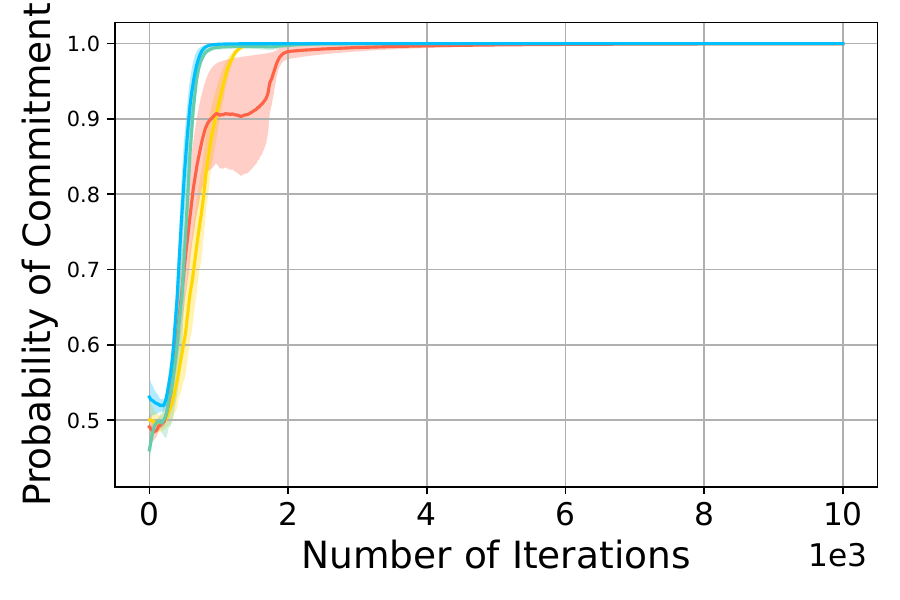}
  \caption{DCL Commitment Policy of $(C,C)$}
  \label{fig:pd-pi_c}
\end{subfigure}%
\caption{DCL Policies in Prisoner's Dilemma}
\label{fig:pd-policies}
\end{figure*}
We evaluated the performance of DCL focusing on two objectives. First, we investigated DCL's ability to foster cooperative behaviors among agents in challenging mixed-motive tasks. To validate this, we analyzed the behaviors of agents with mutual commitment and without commitment. 
Second, we compared DCL's efficiency against other multi-agent reinforcement learning algorithms in tabular, repeated, and sequential social dilemmas. We demonstrated improvements in both agents' self-interest optimization and social welfare. 
Additionally, we compared centralized (Algorithm~\ref{alg:dcl_centralized}, Appendix~\ref{Centralized_DCL_details}) and decentralized (Algorithm~\ref{alg:dcl_decentralized}, Appendix~\ref{Decentralized_DCL_details}) versions of DCL. Each algorithm was executed with and without incentive-compatible constraints (denoted as DCL-IC and DCL respectively), to further explore the impact of the constraints introduced in Equation~(\ref{commitment_eq}). 
\vspace*{-0.1cm}
\subsection{Baselines}
\label{section:baseline}
\begin{figure*}[t]
\centering
\begin{subfigure}{1\textwidth}
  \centering
  \includegraphics[width=1\textwidth]{figures/legend.pdf}
\end{subfigure}%
\vspace{0.1mm}
\begin{subfigure}{.32\textwidth}
  \centering
  \includegraphics[width=1\textwidth]{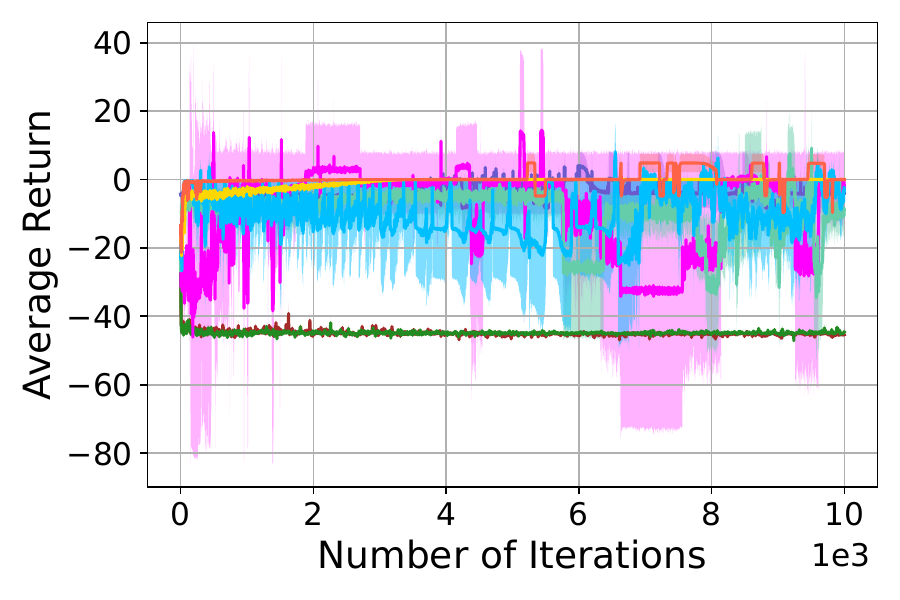}
  \caption{Agent $1$ Return in Grid Game}
  \label{fig:grid-a1}
\end{subfigure}%
\begin{subfigure}{.32\textwidth}
  \centering
  \includegraphics[width=1\textwidth]{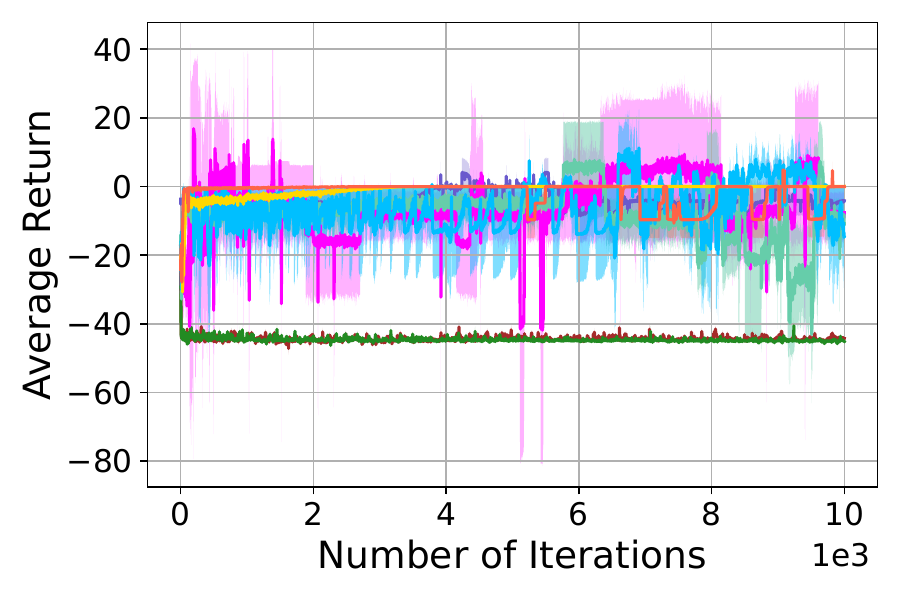}
  \caption{Agent $2$ Return in Grid Game}
  \label{fig:grid-a2}
\end{subfigure}%
\begin{subfigure}{.32\textwidth}
  \centering
  \includegraphics[width=1\textwidth]{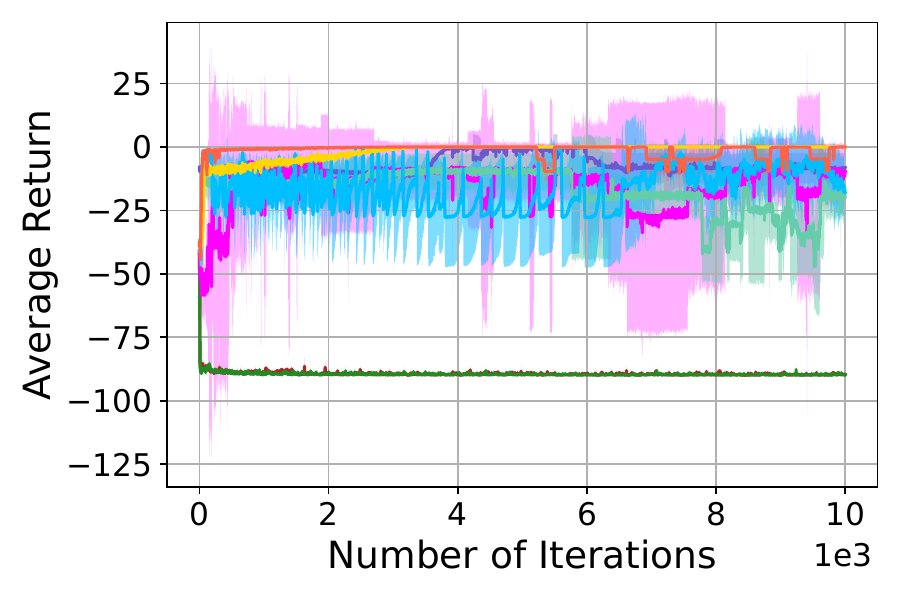}
  \caption{Social Welfare in Grid Game}
  \label{fig:grid-social}
\end{subfigure}%
\caption{Grid Game (Horizon=$16$): DCL v.s. Other Baselines.}
\label{fig:grid-results}
\end{figure*}
\begin{figure*}[htbp]
\centering
\begin{subfigure}{.32\textwidth}
  \centering
  \includegraphics[width=1\textwidth]{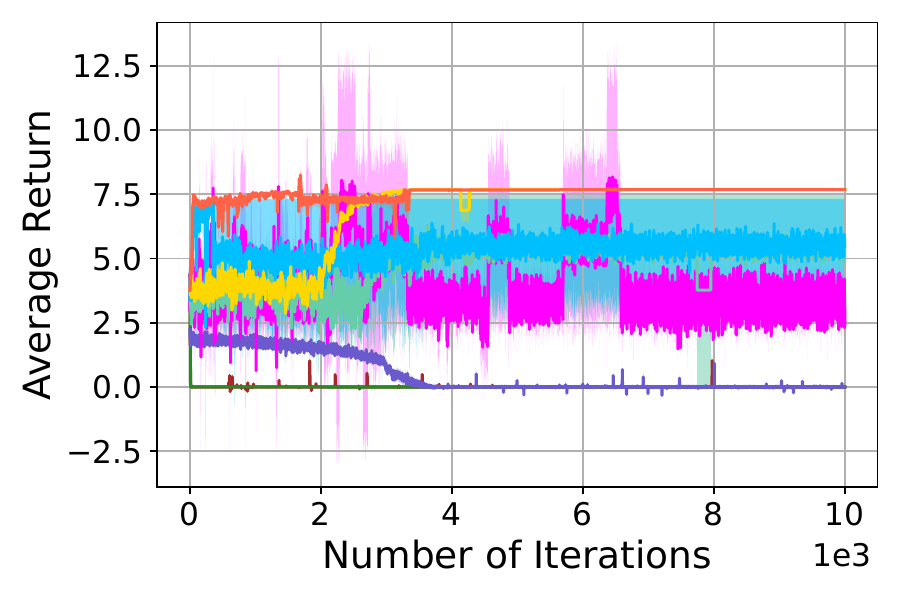}
  \caption{Agent $1$ Return in RPC}
  \label{fig:ipc16-a1}
\end{subfigure}%
\begin{subfigure}{.32\textwidth}
  \centering
  \includegraphics[width=1\textwidth]{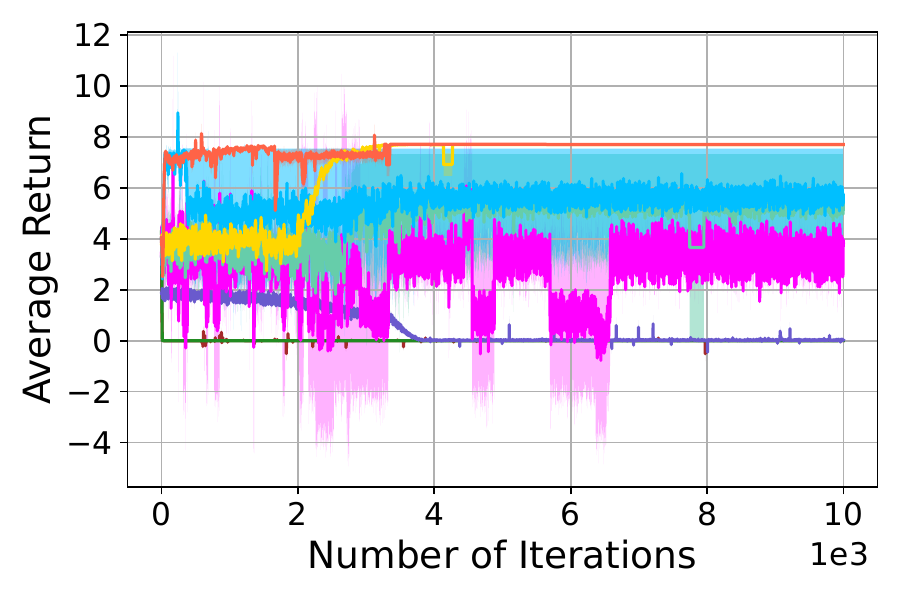}
  \caption{Agent $2$ Return in RPC}
  \label{fig:ipc16-a2}
\end{subfigure}%
\begin{subfigure}{.32\textwidth}
  \centering
  \includegraphics[width=1\textwidth]{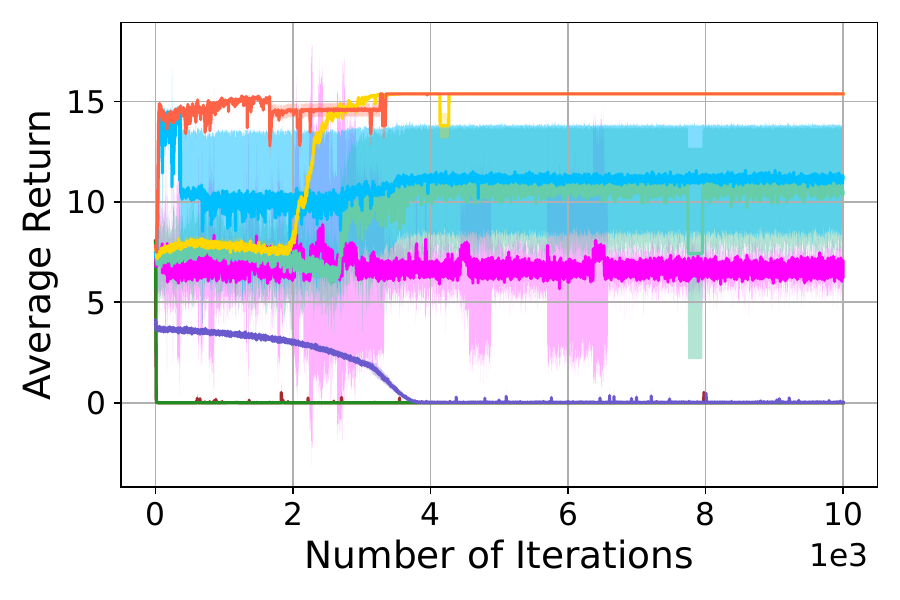}
  \caption{Social Welfare in RPC}
  \label{fig:ipc16-social}
\end{subfigure}%
\caption{Repeated Purely Conflicting Game (Horizon=$16$): DCL v.s. Other Baselines.}
\label{fig:ipc16-results}
\end{figure*}
We compared DCL with the following baselines. Each curve was averaged over 10 seeds with shaded regions indicating standard errors.
Hyperparameters and more implementation details can be found in Appendix~\ref{hyperparam}.
\paragraph{Independent PPO (IPPO)} In this baseline, each agent was trained independently with the proximal policy optimization (PPO)~\citep{schulman2017proximal}. The objective of each agent is maximizing individual expected returns. We implemented multi-agent independent PPO with RLlib~\citep{liang2018rllib}.
\paragraph{Mediated MARL}
To compare with an altruistic third party mechanism, we implemented mediated multi-agent reinforcement learning
using the code released by \citet{ivanov2023mediated}. The mediator, whether constrained or unconstrained, was trained to maximize the utilitarian social welfare, i.e., the expected sum of all agents' returns, while other agents were trained independently to maximize their self-interests. Both agents and the mediator were optimized via actor-critic algorithms~\citep{mnih2016asynchronous}.
\paragraph{Multi-Objective Contract Augmentation Learning (MOCA)} 
To compare with a contract mechanism with reward transfer, we implemented multi-objective contract augmentation learning with the code released by~\citet{christoffersen2024mitigating,haupt2022formal}. Each agent was trained to maximize self-interest, with a learnable transfer payment that directly modifies agents' rewards.
\subsection{Results}
\subsubsection{Prisoner's Dilemma}
Prisoner's Dilemma~\citep{rapoport1965prisoner} is a normal form mixed-motive game, with payoff matrix in Table~\ref{PD_matrix}. 
\vspace*{-0.1cm}
\begin{table}[ht]
\caption{Prisoner's Dilemma} \label{PD_matrix}
\begin{center}
\begin{tabular}{c|c|c}
\hline
  & C       & D       \\ \hline
C & (-1,-1) & (-3,0)  \\
D & (0,-3)  & (-2,-2) \\ \hline
\end{tabular}
\end{center}
\end{table}
\vspace*{-0.1cm}
In accord with Proposition~\ref{equilibrium_pd}, Figure~\ref{fig:pd-results} shows that the DCL agents converge to mutual cooperation in the MCG with utilitarian social welfare $-2$. The fully decentralized DCL also converges to mutual cooperation, while having a larger oscillation before convergence (Figure~\ref{fig:pd-results}). 
This behavior is expected since decentralized DCL estimates policies of other agents rather than directly accessing the true policies, which introduces biases, particularly in the early stages of training. 
These biases are gradually reduced as the estimated policies approach the actual policies over time. 
Figure~\ref{fig:pd-policies} shows the policies of proposals, commitments and actions. Without mutual commitment, the probability of cooperation converges to $0$. Whereas under the conditional commitment mechanism, the probabilities of proposing and committing to mutual cooperation converge to $1$. This result aligns with our theoretical analysis in Proposition~\ref{pd_proof} and demonstrates the capability of commitment mechanism to achieve cooperation.

Mediated MARL with an unconstrained mediator shows the second-best performance, while constrained mediated MARL performs worse, failing to converge to either mutual cooperation or defection. 
This failure may arise from inaccurate value estimation in mediated MARL, which constrains the mediator's policy during training. Specifically, mediated MARL trains each agent with independent actor critic~\citep{mnih2016asynchronous}, considering other agents as part of the environment, leading to nonstationarity from each agent's perspective. In contrast, DCL agents consider joint actions when evaluating future expected returns, avoiding conflicts with the stationary environment assumption in MCGs. Furthermore, the constrained mediated MARL dynamically updates the Lagrange multiplier, shifting the optimization objective at each timestep, which may lead to divergence.

The other baselines, MOCA and IPPO, converge to the mutual defection equilibrium after only a few iterations. Without mechanism design, mutual defection is the only Nash equilibrium in Prisoner's Dilemma, so it is expected that IPPO fails to achieve cooperation.
Without a specific choice of contract space and hand-crafted rules, MOCA also fails to find a contract acceptable to all agents.
\subsubsection{Grid Game}
The above results show that DCL works well on a tabular social dilemma with a single state, we next extend the evaluation to sequential social dilemmas. We created a $2$-player, $T$-step, $N$-grid game, where agent $1$ starts at grid position $p^1_0 = 0$, and agent $2$ starts at $p^2_0 = N - 1$. At each timestep, each player observes both agents' locations, $s_t = (p^1_t, p^2_t)$, and chooses between moving forward, $p^i_{t+1} = \min\{p^i_t + 1, N - 1\}$, or moving backward, $p^i_{t+1} = \max\{p^i_t - 1, 0\}$.
Rewards are defined based on agents' positions: for agent $1$, $r^1 = p^1 - 2(N - 1 - p^2)$; for agent $2$, $r^2 = N - 1 - p^2 - 2p^1$.
This grid game presents a social dilemma at every state. Agents benefit from cooperation by moving away from the other player's initial position, while the dominant strategy is to move towards the other's starting point. Figure~\ref{fig:grid-results} demonstrates that DCL agents gradually learn to cooperate, with zero accumulated discounted rewards. In contrast, other baselines fail to converge to such cooperative strategies.
\subsubsection{Repeated Purely Conflicting Game}
To investigate whether DCL can adapt effectively to scenarios with significant competition, we then introduced a purely conflicting game presented in Table~\ref{IPC_matrix}. In this game, an increase in one agent's payoff always results in a decrease in the payoff of others. The dominant strategy of each agent is to play $A_2$ regardless of the opponent's action, which also holds true in finitely repeated versions (denoted as RPC). Under such conditions, agents have no opportunity to establish $1$-step mutually beneficial agreements. As a result, all players receive zero payoff throughout episodes. 
\vspace*{-0.2cm}
\begin{table}[ht]
\caption{Purely Conflicting Game} \label{IPC_matrix}
\begin{center}
\begin{tabular}{c|c|c}
\hline
  & $A_1$       & $A_2$       \\ \hline
$A_1$ & (0,0) & (-1,2)  \\
$A_2$ & (2,-1)  & (0,0) \\ \hline
\end{tabular}
\end{center}
\end{table}

However, if agents can commit to actions over multiple steps, both can achieve positive long-term returns by committing to a tit-for-tat agreement.
To explore this, we extended DCL with mega-step commitments, enabling agents to commit to multi-step, mutually beneficial proposals. 
Our experiments show that DCL agents successfully converge to cooperative strategies $[(A_1,A_2),(A_2,A_1),...]$ by alternating between $A_1$ and $A_2$ in multiple steps. 
While DCL agents make sacrifices at certain steps, they achieve significantly higher cumulative payoffs over the long run compared to other baselines (Figure~\ref{fig:ipc16-results}), demonstrating DCL's adaptability to highly competitive environments. 
\begin{figure*}[ht]
\centering
\begin{subfigure}{.6\textwidth}
  \centering
  \includegraphics[width=1\textwidth]{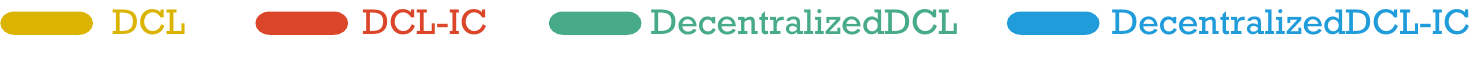}
\end{subfigure}%
\vspace{0.1mm}
\begin{subfigure}{.32\textwidth}
  \centering
  \includegraphics[width=.98\textwidth]{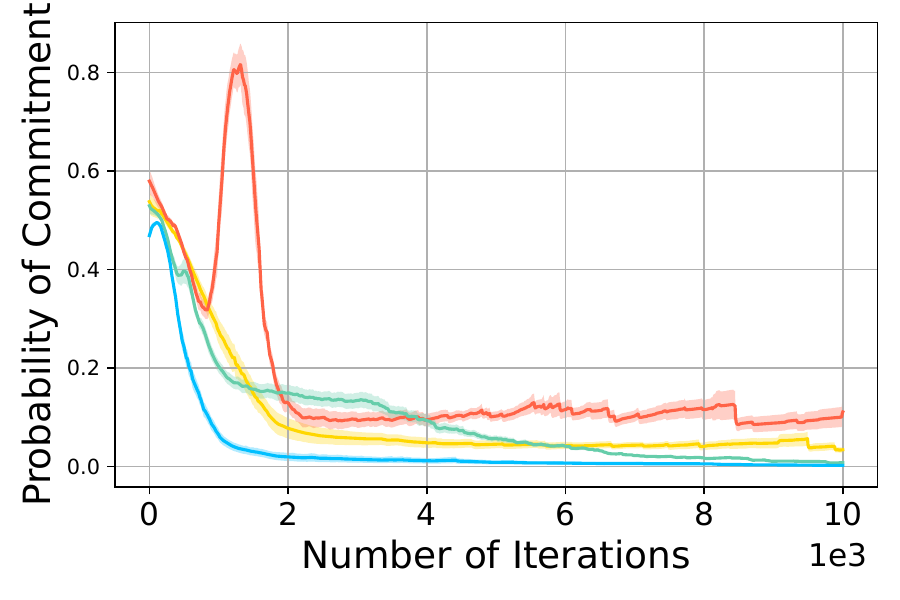}
  \caption{Commitment Policy of $(C,D)$}
  \label{fig:pd-psi-cd}
\end{subfigure}%
\begin{subfigure}{.32\textwidth}
  \centering
  \includegraphics[width=.98\textwidth]{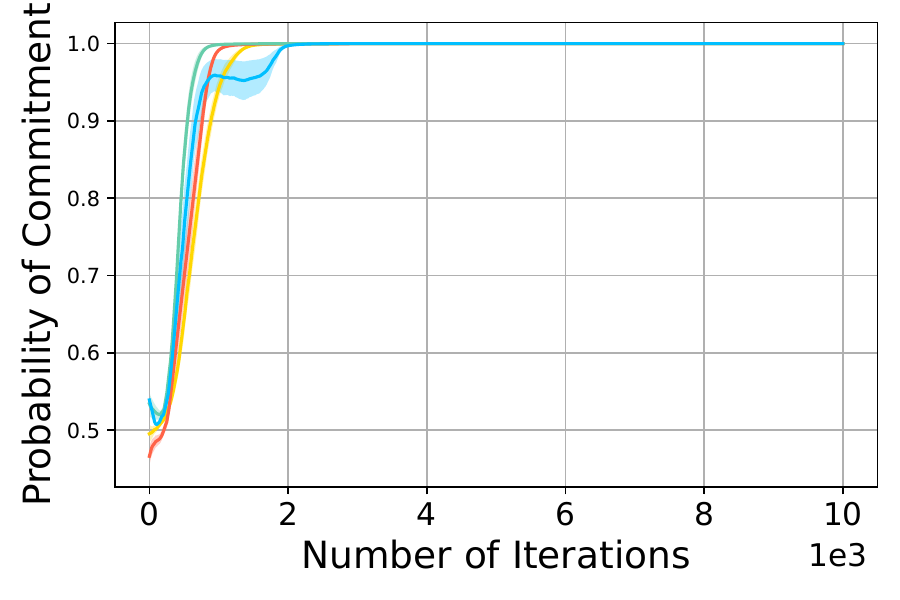}
  \caption{Commitment Policy of $(D,C)$}
  \label{fig:pd-psi-dc}
\end{subfigure}%
\begin{subfigure}{.32\textwidth}
  \centering
  \includegraphics[width=.98\textwidth]{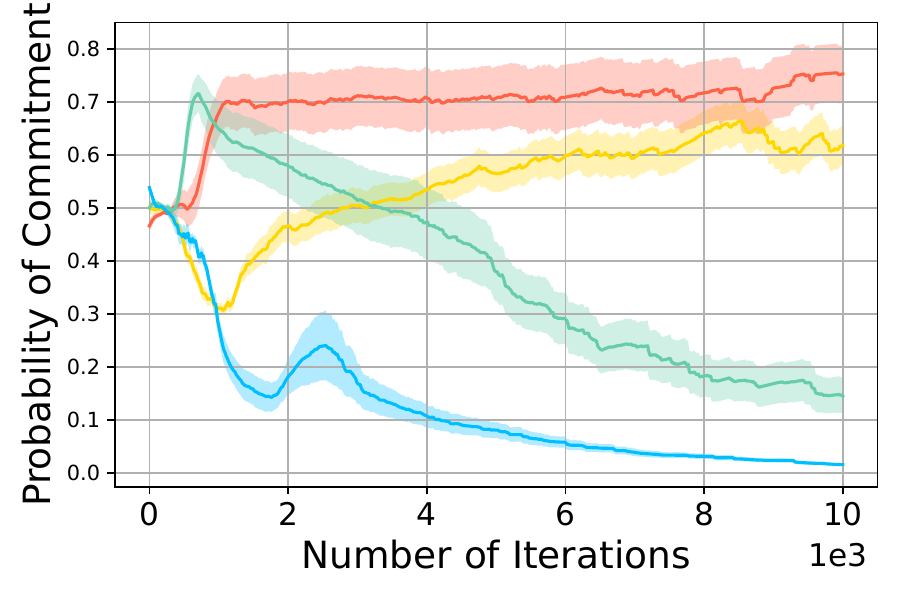}
  \caption{Commitment Policy of $(D,D)$}
  \label{fig:pd-psi-dd}
\end{subfigure}%
\caption{DCL Commitment Policies in Prisoner's Dilemma}
\label{fig:pd-commitment-policies}
\end{figure*}
\section{Discussion on Experiments}
\subsection{Many-player Scenarios}
In MCGs, the joint proposal space grows exponentially with the number of agents, which would inevitably increase the computational complexity. To investigate how DCL handles scalability with many players, we conducted additional experiments on an $N$-player public goods game~\citep{marwell1981economists} with benefit factor $1.5$, where the dominant strategy for each agent is to free-ride by not contributing to the public pool. The results demonstrate that DCL with incentive-compatible constraints performs effectively across scenarios with $2$, $3$, $5$, and $10$ agents, achieving high social welfare. Most agents converge to propose contributions and commit to joint proposals that result in positive individual welfare. These findings indicate that DCL scales well to many-player games, with the agreement rate of joint proposals remaining stable ($>0.99$) as the number of agents increases. We report runtime, average joint proposal agreement rate and average social welfare per batch (batch size =$256$) across $5$ random seeds in Table~\ref{NPD},  Appendix~\ref{additional_exps}.
\subsection{Robustness to Maliciously Irrational Agents}
As shown in Figure~\ref{fig:pd-pi_c} and Figure~\ref{fig:pd-commitment-policies}, DCL agents converge to commitment policies that accept proposals for mutual cooperation and self-defection when the co-player cooperates, while rejecting cooperation when the co-player proposes defection in the Prisoner's Dilemma. Consequently, when interacting with irrational agents—such as those who always propose defection—DCL agents will reject such proposals and choose to defect following their action policies (Figure~\ref{fig:pd-pi_a}). This demonstrates the robustness of DCL agents against malicious agents, as they effectively reject disadvantageous agreements and act in their own best interests.
\section{Conclusion}
We introduced the Markov Commitment Games, a framework that allows self-interested agents to negotiate future plans through voluntary commitments.
It responds to the open problem in cooperative AI~\citep{dafoe2020open} on commitment capabilities without relying on altruism. 
We derived unbiased proposal, commitment, and action policy gradients (Lemma~\ref{gradient_derivation_lemma}), which facilitates the design of policy updates while preserving the stationarity assumption of the multi-agent environment. 
Under the framework of MCGs, we proposed differentiable commitment learning (DCL), which maximizes agents' expected self-interests while incorporating incentive-compatible constraints on their proposal policies to encourage mutually beneficial agreements.
DCL also mitigates limitations of non-stationary training of existing methods. Rather than treating other agents as part of a stationary environment—a simplification that does not hold in multi-agent settings—DCL explicitly leverages other agents' actions when estimating future expected values. 
This approach enhances the accuracy of value estimations and promotes stability during training.
We empirically showed that our method outperforms the baseline methods in multiple tasks, often by successfully facilitating cooperation among agents. We also demonstrated the efficacy of DCL in its fully decentralized implementation. 
\section{Limitations and Future Work}
\paragraph{Sample Efficiency}
Both centralized and decentralized versions of DCL employ on-policy updates for agents' actors and critics, which explores by sampling actions according to the current policy models. This is less sample efficient compared to off-policy methods, which use past trajectories from a replay buffer for model updates. However, off-policy methods may bring biases due to discrepancies between the behavior policy and the target policy. 
While importance sampling can mitigate this issue by re-weighting experiences, it may also introduce high variance, especially when policies diverge significantly. 
Furthermore, in fully decentralized DCL, agents do not have access to other agents' policies, and importance sampling based on estimations of other agents' policies may introduce additional biases.
Therefore, the trade-off between the sample efficiency, bias and variance can be further explored in our future work.
\paragraph{Complex Proposal Domain}
In DCL and MCGs, the proposal domain is formulated as a set of future actions. This reflects real-world scenarios, where agreements often specify future actions conditioned on the behavior of other parties. Nevertheless, human commitments can take various forms, such as stochastic policies of future plan. Extending our framework to accommodate more complex proposal domains presents a promising direction for future research.
\subsubsection*{Acknowledgements}
We acknowledge funding from the 
Canada CIFAR AI Chair program, a discovery grant from the Natural Sciences and Engineering Research Council of Canada and a grant from IITP \& MSIT of Korea (No. RS-2024-00457882, AI Research Hub Project). 
 Computational resources used in preparing this research were provided, in
part, by the Province of Ontario, the Government of Canada
through CIFAR, and companies sponsoring the Vector Institute \url{https://vectorinstitute.ai/partners/}. Baoxiang Wang was sponsored by the Vector Institute's visiting researcher program.
\bibliography{Reference}

\onecolumn
\aistatstitle{Learning to Negotiate via Voluntary Commitment: \\Supplementary Materials}
\appendix
\section{Proof of Lemma~\ref{gradient_derivation_lemma}}
\label{gradient_derive}
The proof of Lemma~\ref{gradient_derivation_lemma} derives the action, commitment, and proposal policy gradients in DCL. Recall that the state value function (the objective function of self-interested agents) in MCGs is:
\begin{equation}
    V^i_{\bm{\phi,\psi,\pi}}(s)=\mathbb{E}_{\bm{\phi,\psi,\pi}}[\sum_{k=t}^\infty \gamma^{k-t} r_{k+1}^i|s_t=s].
\end{equation}
The state-action value function is:
\begin{equation}
    Q^i_{\bm{\phi,\psi,\pi}}(s,\mathbf{a})=\mathbb{E}_{\bm{\phi,\psi,\pi}}[\sum_{k=t}^\infty \gamma^{k-t} r_{k+1}^i|s_t=s,\mathbf{a}_t=\mathbf{a}].
\end{equation}
Therefore we can expand the state value function by:
\begin{equation}
\label{expand_v}
\begin{aligned}
    V^i_{\bm{\phi,\psi,\pi}}(s) =& \sum_{\mathbf{m}\sim \bm{\phi}}\bm{\phi}(\mathbf{m}|s)\sum_{\mathbf{c}\sim \bm{\psi}}\bm{\psi}(\mathbf{c}|s, \mathbf{m})
    \sum_{\mathbf{a} \sim \bm{\pi}}\bm{\pi}(\mathbf{a}|s)
    \Bigg[\mathds{1}(\mathbf{c=1}) Q_{\bm{\phi,\psi,\pi}}^i(s,\mathbf{m})
    +\Big(1-\mathds{1}(\mathbf{c=1})\Big)Q_{\bm{\phi,\psi,\pi}}^i(s,\mathbf{a}) \Bigg].
\end{aligned}
\end{equation}
We then derive policy gradients based on the state-action value function and policy functions.
\subsection{Unconstrained Policy Gradient}
\begin{proof}
First, we consider the action policy gradient $\nabla_{\theta^i} V^i_{\bm{\phi,\psi,\pi}}(s)$ for each agent $i\in \mathcal{N}$:
\begin{equation}
\begin{aligned}
    &\nabla_{\theta^i} V^i_{\bm{\phi,\psi,\pi}}(s)\\
    =& \sum_{\mathbf{m}\sim \bm{\phi}}\bm{\phi}(\mathbf{m}|s)\sum_{\mathbf{c}\sim \bm{\psi}}\bm{\psi}(\mathbf{c}|s, \mathbf{m})\Bigg[\mathds{1}(\mathbf{c=1})\nabla_{\theta^i} Q_{\bm{\phi,\psi,\pi}}^i(s,\mathbf{m})+\Big(1-\mathds{1}(\mathbf{c=1})\Big)\sum_{\mathbf{a} \sim \bm{\pi}}Q_{\bm{\phi,\psi,\pi}}^i(s,\mathbf{a})\nabla_{\theta^i}\bm{\pi}(\mathbf{a}|s)\\
    &+\bm{\pi}(\mathbf{a}|s)\nabla_{\theta^i}Q_{\bm{\phi,\psi,\pi}}^i(s,\mathbf{a}) \Bigg],\\
    =& \sum_{\mathbf{m}\sim \bm{\phi}}\bm{\phi}(\mathbf{m}|s)\sum_{\mathbf{c}\sim \bm{\psi}}\bm{\psi}(\mathbf{c}|s, \mathbf{m})\Bigg[\Big(1-\mathds{1}(\mathbf{c=1})\Big)\sum_{\mathbf{a} \sim \bm{\pi}}Q_{\bm{\phi,\psi,\pi}}^i(s,\mathbf{a})\nabla_{\theta^i}\bm{\pi}(\mathbf{a}|s)\Bigg]+\sum_{\mathbf{m}\sim \bm{\phi}}\bm{\phi}(\mathbf{m}|s)\sum_{\mathbf{c}\sim \bm{\psi}}\bm{\psi}(\mathbf{c}|s, \mathbf{m})\\
    &\cdot \Bigg[\mathds{1}(\mathbf{c=1})\nabla_{\theta^i} Q_{\bm{\phi,\psi,\pi}}^i(s,\mathbf{m})+\Big(1-\mathds{1}(\mathbf{c=1})\Big)\sum_{\mathbf{a} \sim \bm{\pi}}\bm{\pi}(\mathbf{a}|s)\nabla_{\theta^i}Q_{\bm{\phi,\psi,\pi}}^i(s,\mathbf{a}) \Bigg].\\
\end{aligned}
\end{equation}
Let  $f_{\bm{\phi,\psi,\pi}}(s)=\sum_{\mathbf{m}\sim \bm{\phi}}\bm{\phi}(\mathbf{m}|s)\sum_{\mathbf{c}\sim \bm{\psi}}\bm{\psi}(\mathbf{c}|s, \mathbf{m})\Bigg[\Big(1-\mathds{1}(\mathbf{c=1})\Big)\sum_{\mathbf{a} \sim \bm{\pi}}Q_{\bm{\phi,\psi,\pi}}^i(s,\mathbf{a})\nabla_{\theta^i}\bm{\pi}(\mathbf{a}|s)\Bigg]$.
We have
\begin{equation}
\begin{aligned}
    \nabla_{\theta^i} V^i_{\bm{\phi,\psi,\pi}}(s)=&f_{\bm{\phi,\psi,\pi}}(s) + \sum_{\mathbf{m}\sim \bm{\phi}}\bm{\phi}(\mathbf{m}|s)\sum_{\mathbf{c}\sim \bm{\psi}}\bm{\psi}(\mathbf{c}|s, \mathbf{m})\Bigg[\mathds{1}(\mathbf{c=1})\nabla_{\theta^i} Q_{\bm{\phi,\psi,\pi}}^i(s,\mathbf{m})+\Big(1-\mathds{1}(\mathbf{c=1})\Big)\\
    &\cdot \sum_{\mathbf{a} \sim \bm{\pi}}\bm{\pi}(\mathbf{a}|s)\nabla_{\theta^i}Q_{\bm{\phi,\psi,\pi}}^i(s,\mathbf{a}) \Bigg].
\end{aligned}
\end{equation}
Since $Q_{\bm{\phi,\psi,\pi}}^i(s,\mathbf{a})=R^i(s,\mathbf{a})+\gamma \sum_{s'}p(s'|s,\mathbf{a})V_{\bm{\phi,\psi,\pi}}^i(s')$, we obtain
\newpage
\begin{equation}
\begin{aligned}
\nabla_{\theta^i}Q_{\bm{\phi,\psi,\pi}}^i(s,\mathbf{a}) = \nabla_{\theta^i}\Big( R^i(s,\mathbf{a})+\gamma \sum_{s'}p(s'|s,\mathbf{a})V_{\bm{\phi,\psi,\pi}}^i(s')\Big) 
= \gamma \sum_{s'}p(s'|s,\mathbf{a}) \nabla_{\theta^i} V_{\bm{\phi,\psi,\pi}}^i(s').
\end{aligned}
\end{equation}
Therefore,
\begin{equation}
\begin{aligned}
    &\nabla_{\theta^i} V^i_{\bm{\phi,\psi,\pi}}(s)\\
    =& f_{\bm{\phi,\psi,\pi}}(s)+ \gamma \sum_{\mathbf{m}\sim \bm{\phi}}\bm{\phi}(\mathbf{m}|s)\sum_{\mathbf{c}\sim \bm{\psi}}\bm{\psi}(\mathbf{c}|s, \mathbf{m})\Bigg[\mathds{1}(\mathbf{c=1}) \sum_{s'}p(s'|s,\mathbf{m}) \nabla_{\theta^i} V_{\bm{\phi,\psi,\pi}}^i(s')+\Big(1-\mathds{1}(\mathbf{c=1})\Big)\\
    &\cdot \sum_{\mathbf{a} \sim \bm{\pi}}\bm{\pi}(\mathbf{a}|s) \sum_{s'}p(s'|s,\mathbf{a}) \nabla_{\theta^i} V_{\bm{\phi,\psi,\pi}}^i(s') \Bigg].
\end{aligned}\end{equation}
Define $d_{\bm{\phi,\psi,\pi}}(s,s',k)$ as the probability of transitioning from state $s$ to state $s'$ in $k$ steps under $\bm{\phi,\psi,\pi}$, then we have 
\begin{equation}
\label{occupancy_measure}
d_{\bm{\phi,\psi,\pi}}(s,s',1) = \sum_{\mathbf{m}\sim \bm{\phi}}\bm{\phi}(\mathbf{m}|s)\sum_{\mathbf{c}\sim \bm{\psi}}\bm{\psi}(\mathbf{c}|s, \mathbf{m})\Bigg[\mathds{1}(\mathbf{c=1}) p(s'|s,\mathbf{m}) +\Big(1-\mathds{1}(\mathbf{c=1})\Big)\sum_{\mathbf{a} \sim \bm{\pi}}\bm{\pi}(\mathbf{a}|s) p(s'|s,\mathbf{a}) \Bigg],
\end{equation}
and 
\begin{equation}
    d_{\bm{\phi,\psi,\pi}}(s,s',k+1) = \sum_{x}d_{\bm{\phi,\psi,\pi}}(s,x,k)d_{\bm{\phi,\psi,\pi}}(x,s',1).
\end{equation}
Note
\begin{equation}
    d_{\bm{\phi,\psi,\pi}}(s,s,0)=\sum_{x} d_{\bm{\phi,\psi,\pi}}(s,x,0) = 1.
\end{equation}
Then,
\begin{equation}
\begin{aligned}
    &\nabla_{\theta^i} V^i_{\bm{\phi,\psi,\pi}}(s)\\
    =& f_{\bm{\phi,\psi,\pi}}(s)+ \gamma \sum_{s'} \sum_{\mathbf{m}\sim \bm{\phi}}\bm{\phi}(\mathbf{m}|s)\sum_{\mathbf{c}\sim \bm{\psi}}\bm{\psi}(\mathbf{c}|s, \mathbf{m})\Bigg[\mathds{1}(\mathbf{c=1}) p(s'|s,\mathbf{m})+\Big(1-\mathds{1}(\mathbf{c=1})\Big)\sum_{\mathbf{a} \sim \bm{\pi}}\bm{\pi}(\mathbf{a}|s) p(s'|s,\mathbf{a}) \Bigg]\\
    &\cdot \nabla_{\theta^i} V_{\bm{\phi,\psi,\pi}}^i(s'),\\
    =& f_{\bm{\phi,\psi,\pi}}(s)+ \gamma \sum_{s'}d_{\bm{\phi,\psi,\pi}}(s,s',1)\nabla_{\theta^i} V_{\bm{\phi,\psi,\pi}}^i(s').
\end{aligned}
\end{equation}
By induction,
\begin{equation}
\begin{aligned}
    &\nabla_{\theta^i} V^i_{\bm{\phi,\psi,\pi}}(s)\\
    =& f_{\bm{\phi,\psi,\pi}}(s)+ \gamma \sum_{s'}d_{\bm{\phi,\psi,\pi}}(s,s',1) \Big(f_{\bm{\phi,\psi,\pi}}(s')+\gamma \sum_{s''}d_{\bm{\phi,\psi,\pi}}(s',s'',1)\nabla_{\theta^i} V_{\bm{\phi,\psi,\pi}}^i(s'') \Big),\\
    =& f_{\bm{\phi,\psi,\pi}}(s) + \gamma \sum_{s'}d_{\bm{\phi,\psi,\pi}}(s,s',1)f_{\bm{\phi,\psi,\pi}}(s')+\gamma^2\sum_{s''} d_{\bm{\phi,\psi,\pi}}(s,s'',2)\nabla_{\theta^i} V_{\bm{\phi,\psi,\pi}}^i(s''),\\
    =& \sum_{x\in \mathcal{S}}\sum_{k=0}^\infty \gamma^k d_{\bm{\phi,\psi,\pi}}(s,x,k)f_{\bm{\phi,\psi,\pi}}(x).
\end{aligned}
\end{equation}
Then we define a stationary distribution $\rho_{\bm{\phi,\psi,\pi}}(x) = \frac{\sum_{k=0}^\infty \gamma^k d_{\bm{\phi,\psi,\pi}}(s,x,k)}{\sum_{x\in\mathcal{S}}\sum_{k=0}^\infty \gamma^k d_{\bm{\phi,\psi,\pi}}(s,x,k)}$, also known as an occupancy measure of $\bm{\phi,\psi,\pi}$. Thus,
\begin{equation}
\begin{aligned}
    &\nabla_{\theta^i} V^i_{\bm{\phi,\psi,\pi}}(s)\\
    \propto& \sum_{x\in \mathcal{S}} \rho_{\bm{\phi,\psi,\pi}}(x)f_{\bm{\phi,\psi,\pi}}(x),\\
    =& \sum_{x\in \mathcal{S}} \rho_{\bm{\phi,\psi,\pi}}(x)\sum_{\mathbf{m}\sim \bm{\phi}}\bm{\phi}(\mathbf{m}|x)\sum_{\mathbf{c}\sim \bm{\psi}}\bm{\psi}(\mathbf{c}|x, \mathbf{m})\Bigg[\Big(1-\mathds{1}(\mathbf{c=1})\Big)\sum_{\mathbf{a}\sim \bm{\pi}}Q_{\bm{\phi,\psi,\pi}}^i(x,\mathbf{a})\nabla_{\theta^i}\bm{\pi}(\mathbf{a}|x)\Bigg],\\
    =& \sum_{x\in \mathcal{S}} \rho_{\bm{\phi,\psi,\pi}}(x)\sum_{\mathbf{m}\sim \bm{\phi}}\bm{\phi}(\mathbf{m}|x)\sum_{\mathbf{c}\sim \bm{\psi}}\bm{\psi}(\mathbf{c}|x, \mathbf{m}) \sum_{\mathbf{a}\sim \bm{\pi}}\bm{\pi}(\mathbf{a}|x)\Bigg[\Big(1-\mathds{1}(\mathbf{c=1})\Big)Q_{\bm{\phi,\psi,\pi}}^i(x,\mathbf{a})\nabla_{\theta^i}\log \bm{\pi}(\mathbf{a}|x)\Bigg],\\
    =& \mathbb{E}_{x \sim \rho_{\bm{\phi,\psi,\pi}}, \mathbf{m} \sim \bm{\phi}, \mathbf{c} \sim \bm{\psi}, \mathbf{a}\sim\bm{\pi}}\Big[\Big(1-\mathds{1}(\mathbf{c=1})\Big)Q_{\bm{\phi,\psi,\pi}}^i(x,\mathbf{a})\nabla_{\theta^i}\log \bm{\pi}(\mathbf{a}|x) \Big],\\
    =& \mathbb{E}_{x \sim \rho_{\bm{\phi,\psi,\pi}}, \mathbf{m} \sim \bm{\phi}, \mathbf{c} \sim \bm{\psi}, \mathbf{a}\sim\bm{\pi}}\Big[\Big(1-\mathds{1}(\mathbf{c=1})\Big)Q_{\bm{\phi,\psi,\pi}}^i(x,\mathbf{a})\nabla_{\theta^i}\log \pi^i(a^i|x) \Big].
\end{aligned}
\end{equation}
Therefore, we have 
\begin{equation}
    \nabla_{\theta^i} V^i_{\bm{\phi,\psi,\pi}}(s) \propto \mathbb{E}_{x \sim \rho_{\bm{\phi,\psi,\pi}}, \mathbf{m} \sim \bm{\phi}, \mathbf{c} \sim \bm{\psi}, \mathbf{a}\sim\bm{\pi}}\Big[\Big(1-\mathds{1}(\mathbf{c=1})\Big)Q_{\bm{\phi,\psi,\pi}}^i(x,\mathbf{a})\nabla_{\theta^i}\log \pi^i(a^i|x) \Big].\tag*{\qedhere}
\end{equation}
\end{proof}
\subsection{Commitment Network Gradient}
\begin{proof}
Next, we consider commitment policy gradient $\nabla_{\zeta^i} V^i_{\bm{\phi,\psi,\pi}}(s)$:
\begin{equation}
    \begin{aligned}
        &\nabla_{\zeta^i} V^i_{\bm{\phi,\psi,\pi}}(s)\\
        =&\nabla_{\zeta^i} \sum_{\mathbf{m}\sim \bm{\phi}}\bm{\phi}(\mathbf{m}|s)\sum_{\mathbf{c}\sim \bm{\psi}}\bm{\psi}(\mathbf{c}|s, \mathbf{m})\Bigg[\mathds{1}(\mathbf{c=1})Q_{\bm{\phi,\psi,\pi}}^i(s,\mathbf{m})+\Big(1-\mathds{1}(\mathbf{c=1})\Big)\sum_{\mathbf{a} \sim \bm{\pi}}\bm{\pi}(\mathbf{a}|s)Q_{\bm{\phi,\psi,\pi}}^i(s,\mathbf{a}) \Bigg],\\
        =& \sum_{\mathbf{m}\sim \bm{\phi}}\bm{\phi}(\mathbf{m}|s)\sum_{\mathbf{c}\sim \bm{\psi}}\Bigg[\mathds{1}(\mathbf{c=1})Q_{\bm{\phi,\psi,\pi}}^i(s,\mathbf{m})+\Big(1-\mathds{1}(\mathbf{c=1})\Big)\sum_{\mathbf{a} \sim \bm{\pi}}\bm{\pi}(\mathbf{a}|s)Q_{\bm{\phi,\psi,\pi}}^i(s,\mathbf{a}) \Bigg]\nabla_{\zeta^i}\bm{\psi}(\mathbf{c}|s, \mathbf{m})\\
        &+ \sum_{\mathbf{m}\sim \bm{\phi}}\bm{\phi}(\mathbf{m}|s)\sum_{\mathbf{c}\sim \bm{\psi}}\bm{\psi}(\mathbf{c}|s, \mathbf{m})\nabla_{\zeta^i}\Bigg[\mathds{1}(\mathbf{c=1})Q_{\bm{\phi,\psi,\pi}}^i(s,\mathbf{m})+\Big(1-\mathds{1}(\mathbf{c=1})\Big)\sum_{\mathbf{a} \sim \bm{\pi}}\bm{\pi}(\mathbf{a}|s)Q_{\bm{\phi,\psi,\pi}}^i(s,\mathbf{a}) \Bigg],\\
        =& \sum_{\mathbf{m}\sim \bm{\phi}}\bm{\phi}(\mathbf{m}|s)\sum_{\mathbf{c}\sim \bm{\psi}}\Bigg[\mathds{1}(\mathbf{c=1})Q_{\bm{\phi,\psi,\pi}}^i(s,\mathbf{m})+\Big(1-\mathds{1}(\mathbf{c=1})\Big)\sum_{\mathbf{a} \sim \bm{\pi}}\bm{\pi}(\mathbf{a}|s)Q_{\bm{\phi,\psi,\pi}}^i(s,\mathbf{a}) \Bigg]\nabla_{\zeta^i}\bm{\psi}(\mathbf{c}|s, \mathbf{m})\\
        &+ \sum_{\mathbf{m}\sim \bm{\phi}}\bm{\phi}(\mathbf{m}|s)\sum_{\mathbf{c}\sim \bm{\psi}}\bm{\psi}(\mathbf{c}|s, \mathbf{m})\Bigg[Q_{\bm{\phi,\psi,\pi}}^i(s,\mathbf{m})-\sum_{\mathbf{a} \sim \bm{\pi}}\bm{\pi}(\mathbf{a}|s)Q_{\bm{\phi,\psi,\pi}}^i(s,\mathbf{a}) \Bigg]\nabla_{\zeta^i}\mathds{1}(\mathbf{c=1})\\
        &+ \sum_{\mathbf{m}\sim \bm{\phi}}\bm{\phi}(\mathbf{m}|s)\sum_{\mathbf{c}\sim \bm{\psi}}\bm{\psi}(\mathbf{c}|s, \mathbf{m})\Bigg[\mathds{1}(\mathbf{c=1})\nabla_{\zeta^i}Q_{\bm{\phi,\psi,\pi}}^i(s,\mathbf{m})+\Big(1-\mathds{1}(\mathbf{c=1})\Big)\sum_{\mathbf{a} \sim \bm{\pi}}\bm{\pi}(\mathbf{a}|s)\nabla_{\zeta^i}Q_{\bm{\phi,\psi,\pi}}^i(s,\mathbf{a}) \Bigg].
    \end{aligned}
\end{equation}
Let 
\begin{equation*}
\begin{aligned}
    g_{\bm{\phi,\psi,\pi}}(s)=&\sum_{\mathbf{m}\sim \bm{\phi}}\bm{\phi}(\mathbf{m}|s)\sum_{\mathbf{c}\sim \bm{\psi}}\Bigg[\mathds{1}(\mathbf{c=1})Q_{\bm{\phi,\psi,\pi}}^i(s,\mathbf{m})+\Big(1-\mathds{1}(\mathbf{c=1})\Big)\sum_{\mathbf{a} \sim \bm{\pi}}\bm{\pi}(\mathbf{a}|s)Q_{\bm{\phi,\psi,\pi}}^i(s,\mathbf{a}) \Bigg]\nabla_{\zeta^i}\bm{\psi}(\mathbf{c}|s, \mathbf{m})\\
    &+ \sum_{\mathbf{m}\sim \bm{\phi}}\bm{\phi}(\mathbf{m}|s)\sum_{\mathbf{c}\sim \bm{\psi}}\bm{\psi}(\mathbf{c}|s, \mathbf{m})\Bigg[Q_{\bm{\phi,\psi,\pi}}^i(s,\mathbf{m})-\sum_{\mathbf{a} \sim \bm{\pi}}\bm{\pi}(\mathbf{a}|s)Q_{\bm{\phi,\psi,\pi}}^i(s,\mathbf{a}) \Bigg]\nabla_{\zeta^i}\mathds{1}(\mathbf{c=1}).
\end{aligned}
\end{equation*}
Then,
\begin{equation}
    \begin{aligned}
        &\nabla_{\zeta^i} V^i_{\bm{\phi,\psi,\pi}}(s)\\
        =& g_{\bm{\phi,\psi,\pi}}(s)+ \sum_{\mathbf{m}\sim \bm{\phi}}\bm{\phi}(\mathbf{m}|s)\sum_{\mathbf{c}\sim \bm{\psi}}\bm{\psi}(\mathbf{c}|s, \mathbf{m})\Bigg[\mathds{1}(\mathbf{c=1})\nabla_{\zeta^i}Q_{\bm{\phi,\psi,\pi}}^i(s,\mathbf{m})+\Big(1-\mathds{1}(\mathbf{c=1})\Big)\sum_{\mathbf{a} \sim \bm{\pi}}\bm{\pi}(\mathbf{a}|s)\\
        &\cdot \nabla_{\zeta^i}Q_{\bm{\phi,\psi,\pi}}^i(s,\mathbf{a}) \Bigg],\\
        =& g_{\bm{\phi,\psi,\pi}}(s)+\gamma \sum_{\mathbf{m}\sim \bm{\phi}}\bm{\phi}(\mathbf{m}|s)\sum_{\mathbf{c}\sim \bm{\psi}}\bm{\psi}(\mathbf{c}|s, \mathbf{m})\Bigg[\mathds{1}(\mathbf{c=1})\sum_{s'} p(s'|s,\mathbf{m})\nabla_{\zeta^i}V_{\bm{\phi,\psi,\pi}}^i(s')+\Big(1-\mathds{1}(\mathbf{c=1})\Big)\\
        &\cdot \sum_{\mathbf{a} \sim \bm{\pi}}\bm{\pi}(\mathbf{a}|s)\sum_{s'}p(s'|s,\mathbf{a})\nabla_{\zeta^i}V_{\bm{\phi,\psi,\pi}}^i(s') \Bigg].
    \end{aligned}
\end{equation}
According to \eqref{occupancy_measure}, 
\begin{equation}
        \nabla_{\zeta^i} V^i_{\bm{\phi,\psi,\pi}}(s)= g_{\bm{\phi,\psi,\pi}}(s)+\gamma\sum_{s'}d_{\bm{\phi,\psi,\pi}}(s,s',1)\nabla_{\zeta^i}V_{\bm{\phi,\psi,\pi}}^i(s').
\end{equation}
Similarly by induction,
\begin{equation}
\begin{aligned}
    &\nabla_{\zeta^i} V^i_{\bm{\phi,\psi,\pi}}(s)\\
    =&\sum_{x\in \mathcal{S}}\sum_{k=0}^\infty \gamma^k d_{\bm{\phi,\psi,\pi}}(s,x,k)g_{\bm{\phi,\psi,\pi}}(x),\\
    \propto& \sum_{x\in \mathcal{S}} \rho_{\bm{\phi,\psi,\pi}}(x)g_{\bm{\phi,\psi,\pi}}(x),\\
    =& \sum_{x\in \mathcal{S}} \rho_{\bm{\phi,\psi,\pi}}(x)\Bigg[\sum_{\mathbf{m}\sim \bm{\phi}}\bm{\phi}(\mathbf{m}|x)\sum_{\mathbf{c}\sim \bm{\psi}}\Big[\mathds{1}(\mathbf{c=1})Q_{\bm{\phi,\psi,\pi}}^i(x,\mathbf{m})+\Big(1-\mathds{1}(\mathbf{c=1})\Big)\sum_{\mathbf{a} \sim \bm{\pi}}\bm{\pi}(\mathbf{a}|x)Q_{\bm{\phi,\psi,\pi}}^i(x,\mathbf{a}) \Big]\\
    &\cdot \nabla_{\zeta^i}\bm{\psi}(\mathbf{c}|x, \mathbf{m})+\sum_{\mathbf{m}\sim \bm{\phi}}\bm{\phi}(\mathbf{m}|x)\sum_{\mathbf{c}\sim \bm{\psi}}\bm{\psi}(\mathbf{c}|x, \mathbf{m})\Big[Q_{\bm{\phi,\psi,\pi}}^i(x,\mathbf{m})-\sum_{\mathbf{a} \sim \bm{\pi}}\bm{\pi}(\mathbf{a}|x)Q_{\bm{\phi,\psi,\pi}}^i(x,\mathbf{a}) \Big]\nabla_{\zeta^i}\mathds{1}(\mathbf{c=1})\Bigg],\\
    =& \mathbb{E}_{x \sim \rho_{\bm{\phi,\psi,\pi}}, \mathbf{m} \sim \bm{\phi}, \mathbf{c} \sim \bm{\psi}, \mathbf{a}\sim\bm{\pi}}\Bigg[ \Big[\mathds{1}(\mathbf{c=1})Q_{\bm{\phi,\psi,\pi}}^i(x,\mathbf{m})+\Big(1-\mathds{1}(\mathbf{c=1})\Big)Q_{\bm{\phi,\psi,\pi}}^i(x,\mathbf{a})\Big]\nabla_{\zeta^i}\log\bm{\psi}(\mathbf{c}|x, \mathbf{m})\\
    &+ \Big[Q_{\bm{\phi,\psi,\pi}}^i(x,\mathbf{m})-Q_{\bm{\phi,\psi,\pi}}^i(x,\mathbf{a}) \Big]\nabla_{\zeta^i}\mathds{1}(\mathbf{c=1})
    \Bigg],\\
    =& \mathbb{E}_{x \sim \rho_{\bm{\phi,\psi,\pi}}, \mathbf{m} \sim \bm{\phi}, \mathbf{c} \sim \bm{\psi}, \mathbf{a}\sim\bm{\pi}}\Bigg[ \Big[\mathds{1}(\mathbf{c=1})Q_{\bm{\phi,\psi,\pi}}^i(x,\mathbf{m})+\Big(1-\mathds{1}(\mathbf{c=1})\Big)Q_{\bm{\phi,\psi,\pi}}^i(x,\mathbf{a})\Big]\nabla_{\zeta^i}\log\psi^i(c^i|x, \mathbf{m})\\
    &+ \Big[Q_{\bm{\phi,\psi,\pi}}^i(x,\mathbf{m})-Q_{\bm{\phi,\psi,\pi}}^i(x,\mathbf{a}) \Big]\prod_{k \neq i}\mathds{1}(c^{k}=1)\nabla_{\zeta^i}\mathds{1}(c^i=1)
    \Bigg].
\end{aligned}
\end{equation}
Therefore, 
\begin{align*}
    \nabla_{\zeta^i} V^i_{\bm{\phi,\psi,\pi}}(s)
    \propto & \mathbb{E}_{x \sim \rho_{\bm{\phi,\psi,\pi}}, \mathbf{m} \sim \bm{\phi}, \mathbf{c} \sim \bm{\psi}, \mathbf{a}\sim\bm{\pi}}\Bigg[ \Big[\mathds{1}(\mathbf{c=1})Q_{\bm{\phi,\psi,\pi}}^i(x,\mathbf{m})+\Big(1-\mathds{1}(\mathbf{c=1})\Big)Q_{\bm{\phi,\psi,\pi}}^i(x,\mathbf{a})\Big]\\
    &\cdot \nabla_{\zeta^i}\log\psi^i(c^i|x, \mathbf{m})+ \Big[Q_{\bm{\phi,\psi,\pi}}^i(x,\mathbf{m})-Q_{\bm{\phi,\psi,\pi}}^i(x,\mathbf{a}) \Big]\prod_{k \neq i}\mathds{1}(c^{k}=1)\nabla_{\zeta^i}\mathds{1}(c^i=1)
    \Bigg]. \tag*{\qedhere}
\end{align*}
\end{proof}
Note that $\nabla_{\zeta^i}\mathds{1}(c^i=1) = \frac{d\mathds{1}(c^i=1)}{dc^i}\frac{\partial c^i}{\partial \zeta^i}$. To compute $\frac{\partial c^i}{\partial \zeta^i}$, we apply the Gumbel-Softmax distribution~\citep{jang2016categorical} for differentiable sampling. This allows backpropagation through the differentiable commitment sample $c^i$ for $\forall i\in \mathcal{N}$.
\subsection{Proposing Network Gradient}
\begin{proof}
Finally, we consider the proposal policy gradient $\nabla_{\eta^i} V^i_{\bm{\phi,\psi,\pi}}(s)$:
\begin{equation}
    \begin{aligned}
        &\nabla_{\eta^i} V^i_{\bm{\phi,\psi,\pi}}(s)\\
        =& \nabla_{\eta^i} \sum_{\mathbf{m}\sim \bm{\phi}}\bm{\phi}(\mathbf{m}|s)\sum_{\mathbf{c}\sim \bm{\psi}}\bm{\psi}(\mathbf{c}|s, \mathbf{m})\Bigg[\mathds{1}(\mathbf{c=1})Q_{\bm{\phi,\psi,\pi}}^i(s,\mathbf{m})+\Big(1-\mathds{1}(\mathbf{c=1})\Big)\sum_{\mathbf{a} \sim \bm{\pi}}\bm{\pi}(\mathbf{a}|s)Q_{\bm{\phi,\psi,\pi}}^i(s,\mathbf{a}) \Bigg],\\
        =& \sum_{\mathbf{m}\sim \bm{\phi}}\sum_{\mathbf{c}\sim \bm{\psi}}\bm{\psi}(\mathbf{c}|s, \mathbf{m})\Bigg[\mathds{1}(\mathbf{c=1})Q_{\bm{\phi,\psi,\pi}}^i(s,\mathbf{m})+\Big(1-\mathds{1}(\mathbf{c=1})\Big)\sum_{\mathbf{a} \sim \bm{\pi}}\bm{\pi}(\mathbf{a}|s)Q_{\bm{\phi,\psi,\pi}}^i(s,\mathbf{a}) \Bigg]\nabla_{\eta^i}\bm{\phi}(\mathbf{m}|s)\\
        &+ \sum_{\mathbf{m}\sim \bm{\phi}}\bm{\phi}(\mathbf{m}|s)\sum_{\mathbf{c}\sim \bm{\psi}}\Bigg[\mathds{1}(\mathbf{c=1})Q_{\bm{\phi,\psi,\pi}}^i(s,\mathbf{m})+\Big(1-\mathds{1}(\mathbf{c=1})\Big)\sum_{\mathbf{a} \sim \bm{\pi}}\bm{\pi}(\mathbf{a}|s)Q_{\bm{\phi,\psi,\pi}}^i(s,\mathbf{a}) \Bigg]\nabla_{\eta^i}\bm{\psi}(\mathbf{c}|s, \mathbf{m})\\
        &+ \sum_{\mathbf{m}\sim \bm{\phi}}\bm{\phi}(\mathbf{m}|s)\sum_{\mathbf{c}\sim \bm{\psi}}\bm{\psi}(\mathbf{c}|s, \mathbf{m})\Bigg[Q_{\bm{\phi,\psi,\pi}}^i(s,\mathbf{m})-\sum_{\mathbf{a} \sim \bm{\pi}}\bm{\pi}(\mathbf{a}|s)Q_{\bm{\phi,\psi,\pi}}^i(s,\mathbf{a}) \Bigg]\nabla_{\eta^i}\mathds{1}(\mathbf{c=1})\\
        &+ \sum_{\mathbf{m}\sim \bm{\phi}}\bm{\phi}(\mathbf{m}|s)\sum_{\mathbf{c}\sim \bm{\psi}}\bm{\psi}(\mathbf{c}|s, \mathbf{m})\Bigg[ 
        \mathds{1}(\mathbf{c=1})\nabla_{\eta^i}Q_{\bm{\phi,\psi,\pi}}^i(s,\mathbf{m})\\
        &+\Big(1-\mathds{1}(\mathbf{c=1})\Big)\sum_{\mathbf{a} \sim \bm{\pi}}\bm{\pi}(\mathbf{a}|s)\nabla_{\eta^i}Q_{\bm{\phi,\psi,\pi}}^i(s,\mathbf{a})
        \Bigg].
    \end{aligned}
\end{equation}
Let
\begin{equation}
    \begin{aligned}
        &h_{\bm{\phi,\psi,\pi}}(s)\\
        =& \sum_{\mathbf{m}\sim \bm{\phi}}\sum_{\mathbf{c}\sim \bm{\psi}}\bm{\psi}(\mathbf{c}|s, \mathbf{m})\Bigg[\mathds{1}(\mathbf{c=1})Q_{\bm{\phi,\psi,\pi}}^i(s,\mathbf{m})+\Big(1-\mathds{1}(\mathbf{c=1})\Big)\sum_{\mathbf{a} \sim \bm{\pi}}\bm{\pi}(\mathbf{a}|s)Q_{\bm{\phi,\psi,\pi}}^i(s,\mathbf{a}) \Bigg]\nabla_{\eta^i}\bm{\phi}(\mathbf{m}|s)\\
        &+ \sum_{\mathbf{m}\sim \bm{\phi}}\bm{\phi}(\mathbf{m}|s)\sum_{\mathbf{c}\sim \bm{\psi}}\Bigg[\mathds{1}(\mathbf{c=1})Q_{\bm{\phi,\psi,\pi}}^i(s,\mathbf{m})+\Big(1-\mathds{1}(\mathbf{c=1})\Big)\sum_{\mathbf{a} \sim \bm{\pi}}\bm{\pi}(\mathbf{a}|s)Q_{\bm{\phi,\psi,\pi}}^i(s,\mathbf{a}) \Bigg]\nabla_{\eta^i}\bm{\psi}(\mathbf{c}|s, \mathbf{m})\\
        &+ \sum_{\mathbf{m}\sim \bm{\phi}}\bm{\phi}(\mathbf{m}|s)\sum_{\mathbf{c}\sim \bm{\psi}}\bm{\psi}(\mathbf{c}|s, \mathbf{m})\Bigg[Q_{\bm{\phi,\psi,\pi}}^i(s,\mathbf{m})-\sum_{\mathbf{a} \sim \bm{\pi}}\bm{\pi}(\mathbf{a}|s)Q_{\bm{\phi,\psi,\pi}}^i(s,\mathbf{a}) \Bigg]\nabla_{\eta^i}\mathds{1}(\mathbf{c=1}).
    \end{aligned}
\end{equation}
Similarly we have
\allowdisplaybreaks
    \begin{align}
        &\nabla_{\eta^i} V^i_{\bm{\phi,\psi,\pi}}(s) \nonumber\\
        \propto& \sum_{x\in \mathcal{S}} \rho_{\bm{\phi,\psi,\pi}}(x)h_{\bm{\phi,\psi,\pi}}(x),\nonumber\\
        =& \sum_{x\in \mathcal{S}} \rho_{\bm{\phi,\psi,\pi}}(x)\Bigg[
        \sum_{\mathbf{m}\sim \bm{\phi}}\sum_{\mathbf{c}\sim \bm{\psi}}\bm{\psi}(\mathbf{c}|x, \mathbf{m})\Big[\mathds{1}(\mathbf{c=1})Q_{\bm{\phi,\psi,\pi}}^i(x,\mathbf{m})\nonumber+\Big(1-\mathds{1}(\mathbf{c=1})\Big)\sum_{\mathbf{a} \sim \bm{\pi}}\bm{\pi}(\mathbf{a}|x)Q_{\bm{\phi,\psi,\pi}}^i(x,\mathbf{a}) \Big]\nonumber\\
        &\cdot \nabla_{\eta^i}\bm{\phi}(\mathbf{m}|x)+ \sum_{\mathbf{m}\sim \bm{\phi}}\bm{\phi}(\mathbf{m}|x)\sum_{\mathbf{c}\sim \bm{\psi}}\Big[\mathds{1}(\mathbf{c=1})Q_{\bm{\phi,\psi,\pi}}^i(x,\mathbf{m})+\Big(1-\mathds{1}(\mathbf{c=1})\Big)\sum_{\mathbf{a} \sim \bm{\pi}}\bm{\pi}(\mathbf{a}|x)Q_{\bm{\phi,\psi,\pi}}^i(x,\mathbf{a}) \Big]\nonumber\\
        &\cdot \nabla_{\eta^i}\bm{\psi}(\mathbf{c}|x, \mathbf{m})+ \sum_{\mathbf{m}\sim \bm{\phi}}\bm{\phi}(\mathbf{m}|x)\sum_{\mathbf{c}\sim \bm{\psi}}\bm{\psi}(\mathbf{c}|x, \mathbf{m})\Big[Q_{\bm{\phi,\psi,\pi}}^i(x,\mathbf{m})-\sum_{\mathbf{a} \sim \bm{\pi}}\bm{\pi}(\mathbf{a}|x)Q_{\bm{\phi,\psi,\pi}}^i(x,\mathbf{a}) \Big]\nabla_{\eta^i}\mathds{1}(\mathbf{c=1})
        \Bigg],\nonumber\\
        =& \mathbb{E}_{x \sim \rho_{\bm{\phi,\psi,\pi}}, \mathbf{m} \sim \bm{\phi}, \mathbf{c} \sim \bm{\psi}, \mathbf{a}\sim\bm{\pi}}\Bigg[
        \Big[\mathds{1}(\mathbf{c=1})Q_{\bm{\phi,\psi,\pi}}^i(x,\mathbf{m})+\Big(1-\mathds{1}(\mathbf{c=1})\Big)Q_{\bm{\phi,\psi,\pi}}^i(x,\mathbf{a}) \Big]\Big(\nabla_{\eta^i}\log\bm{\phi}(\mathbf{m}|x)\nonumber\\
        &+\nabla_{\eta^i}\log\bm{\psi}(\mathbf{c}|x, \mathbf{m}) \Big)+ \Big[Q_{\bm{\phi,\psi,\pi}}^i(x,\mathbf{m})-Q_{\bm{\phi,\psi,\pi}}^i(x,\mathbf{a}) \Big]\nabla_{\eta^i}\mathds{1}(\mathbf{c=1})
        \Bigg],\nonumber\\
        =& \mathbb{E}_{x \sim \rho_{\bm{\phi,\psi,\pi}}, \mathbf{m} \sim \bm{\phi}, \mathbf{c} \sim \bm{\psi}, \mathbf{a}\sim\bm{\pi}}\Bigg[
        \Big[\mathds{1}(\mathbf{c=1})Q_{\bm{\phi,\psi,\pi}}^i(x,\mathbf{m})+\Big(1-\mathds{1}(\mathbf{c=1})\Big)Q_{\bm{\phi,\psi,\pi}}^i(x,\mathbf{a}) \Big]\Big(\nabla_{\eta^i}\log\phi^i(m^i|x)\nonumber\\
        &+\sum_j\nabla_{\eta^i}\log\psi^j(c^j|x, \mathbf{m}) \Big)+ \Big[Q_{\bm{\phi,\psi,\pi}}^i(x,\mathbf{m})-Q_{\bm{\phi,\psi,\pi}}^i(x,\mathbf{a}) \Big]\Big(\mathds{1}(c^{-i}=1)\nabla_{\eta^i}\mathds{1}(c^i=1)+\mathds{1}(c^{i}=1)\nonumber\\
        &\cdot \nabla_{\eta^i}\mathds{1}(c^{-i}=1)\Big)
        \Bigg], \nonumber\\
        =& \mathbb{E}_{x \sim \rho_{\bm{\phi,\psi,\pi}}, \mathbf{m} \sim \bm{\phi}, \mathbf{c} \sim \bm{\psi}, \mathbf{a}\sim\bm{\pi}}\Bigg[
        \Big[\mathds{1}(\mathbf{c=1})Q_{\bm{\phi,\psi,\pi}}^i(x,\mathbf{m})+\Big(1-\mathds{1}(\mathbf{c=1})\Big)Q_{\bm{\phi,\psi,\pi}}^i(x,\mathbf{a}) \Big]\Big(\nabla_{\eta^i}\log\phi^i(m^i|x)\nonumber\\
        &+\sum_j\nabla_{\eta^i}\log\psi^j(c^j|x, \mathbf{m}) \Big)+ \sum_{j}\prod_{k\neq j} \mathds{1}(c^k=1)\Big[Q_{\bm{\phi,\psi,\pi}}^i(x,\mathbf{m})-Q_{\bm{\phi,\psi,\pi}}^i(x,\mathbf{a}) \Big]\nabla_{\eta^i}\mathds{1}(c^j=1)
        \Bigg].
    \end{align}
Therefore, 
\begin{align}
        \nabla_{\eta^i} V^i_{\bm{\phi,\psi,\pi}}(s)\propto&  \mathbb{E}_{x \sim \rho_{\bm{\phi,\psi,\pi}}, \mathbf{m} \sim \bm{\phi}, \mathbf{c} \sim \bm{\psi}, \mathbf{a}\sim\bm{\pi}}\Bigg[
        \Big[\mathds{1}(\mathbf{c=1})Q_{\bm{\phi,\psi,\pi}}^i(x,\mathbf{m})+\Big(1-\mathds{1}(\mathbf{c=1})\Big)Q_{\bm{\phi,\psi,\pi}}^i(x,\mathbf{a}) \Big] \cdot\nonumber \\
&\Big(\nabla_{\eta^i}\log\phi^i(m^i|x)
        +\sum_j\nabla_{\eta^i}\log\psi^j(c^j|x, \mathbf{m}) \Big)+ \sum_{j}\prod_{k\neq j} \mathds{1}(c^k=1)\Big[Q_{\bm{\phi,\psi,\pi}}^i(x,\mathbf{m})-\nonumber\\
        &Q_{\bm{\phi,\psi,\pi}}^i(x,\mathbf{a}) \Big]\nabla_{\eta^i}\mathds{1}(c^j=1)
        \Bigg]. \tag*{\qedhere}
\end{align}
\end{proof}
Note that $\nabla_{\eta^i}\mathds{1}(c^i=1)=\frac{d \mathds{1}(c^i=1)}{d c^i}(\frac{\partial\psi^i }{\partial c^i})^{-1}\frac{\partial \psi^i}{\partial m^i}\frac{\partial m^i}{\partial \eta^i}$, 
$\nabla_{\eta^i}\mathds{1}(c^j=1)|_{j\neq i} = \frac{d \mathds{1}(c^j=1)}{d c^j}(\frac{\partial\psi^j }{\partial c^j})^{-1}\frac{\partial \psi^j}{\partial m^i}\frac{\partial m^i}{\partial \eta^i}$.
We apply Gumbel-Softmax distribution~\citep{jang2016categorical} again, which allows autodifferentiation through $m^i, \forall i$.
\section{DCL Details}
\label{DCL_details}
\subsection{Centralized DCL}
\label{Centralized_DCL_details}
DCL updates policies with respect to policy gradients in Lemma~\ref{gradient_derivation_lemma}. Because calculating $\nabla_{\eta^i} V^i_{\bm{\phi,\psi,\pi}}(s)$ requires differentiation through commitment policies of other agents $j\in \mathcal{N}\setminus i$, we present centralized DCL in Algorithm~\ref{alg:dcl_centralized} that allows agents to backpropagate through exact policies of others. 
\subsection{Decentralized DCL}
\label{Decentralized_DCL_details}
Centralized training is not always feasible in mixed-motive environments. To address this limitation, we further present decentralized DCL in Algorithm~\ref{alg:dcl_decentralized}. In decentralized DCL, each agent estimates others' policies and value functions with DCL. Then, agents can differentiate through these estimates to update their own policies.
\section{Proof of Proposition~\ref{equilibrium_pd}}
Recall the definition of Nash equilibrium and Pareto-dominant outcome:
\begin{definition}{\citep{hu2003nash}}
    In stochastic game $\Gamma$, a Nash equilibrium point is a tuple of n strategies $(\pi^1_*,...,\pi^n_*)$ such that for all $s \in \mathcal{S}$ and $i=1,...,n,$
    \begin{equation}
    \label{nash_eq}
        V^i(s,\pi^1_*, ..., \pi^n_*) \geq V^i(s, \pi^1_*, ..., \pi^{i-1}_*, \pi^i, \pi^{i+1}_*,..., \pi^n_*),~~~\forall \pi^i \in \Pi^i,
    \end{equation}
where $\Pi^i$ is the set of strategies available to agent $i$.
\end{definition}
At a Nash equilibrium, no player can improve their payoff by changing their strategy, assuming that the other players stick to their current strategies.
\begin{definition}{\citep{censor1977pareto,fudenberg1991game}} 
    An outcome of a game is Pareto-dominant, also known as Pareto-optimal and Pareto-efficient, if it's impossible to make one player better-off, without making some other players worse-off.
\end{definition}
To prove Proposition~\ref{equilibrium_pd}, we need to find a tuple of strategies $((\phi^i_*, \psi^i_*, \pi^i_*),(\phi^{-i}_*, \psi^{-i}_*, \pi^{-i}_*))$ in MCGs that satisfies the conditions of Nash equilibrium and Pareto-optimality.
\begin{proof}
\label{pd_proof}
    In the MCG of the Prisoner's Dilemma, we define a tuple of deterministic strategies $\forall i\in \mathcal{N},$ $\phi^i(s)=C$, $\pi^i_*(s)=D$ and 
    \begin{equation}
    \label{optimal_psi}
    \begin{aligned}
        \psi^i_*(s, \mathbf{m}=\{C,C\})=&1,\\
        \psi^i_*(s, \mathbf{m}=\{D,C\})=&1,\\
        \psi^i_*(s, \mathbf{m}=\{C,D\})=&0,\\
        \psi^i_*(s, \mathbf{m}=\{D,D\})=&0~\text{or}~1.
    \end{aligned}
    \end{equation}
So the value function of this tuple is:
    \begin{equation}
        V^i(s,(\phi^i_*, \psi^i_*, \pi^i_*)|(\phi^{-i}_*, \psi^{-i}_*, \pi^{-i}_*))=-1.
    \end{equation}
Then we show that no player can increase their payoff by unilaterally changing to other deterministic strategies, assuming all other players keep their strategies fixed.
    \begin{enumerate}
        \item $\forall \phi^i\neq\phi^i_*$, i.e. $ \phi^i(s)=D$, and $\forall \psi^i$:
        \begin{enumerate}
            \item if $\pi^i(s)=C$,
            \begin{equation}
            V^i(s,(\phi^i, \psi^i, \pi^i)|(\phi^{-i}_*, \psi^{-i}_*, \pi^{-i}_*)) = -3 < -1 = V^i(s,(\phi^i_*, \psi^i_*, \pi^i_*)|(\phi^{-i}_*, \psi^{-i}_*, \pi^{-i}_*)),
            \end{equation}
            \item otherwise $\pi^i(s)=D$,
            \begin{equation}
            V^i(s,(\phi^i, \psi^i, \pi^i)|(\phi^{-i}_*, \psi^{-i}_*, \pi^{-i}_*)) = -2 < -1 = V^i(s,(\phi^i_*, \psi^i_*, \pi^i_*)|(\phi^{-i}_*, \psi^{-i}_*, \pi^{-i}_*)).
        \end{equation}
        \end{enumerate}

        \item $\phi^i =\phi^i_*$ and $\psi^i\neq \psi^i_*$:
        \begin{enumerate}
            \item $\forall \psi^i\neq \psi^i_*$ s.t. $\psi^i(s, \mathbf{m}=\{C,C\})=1$ and $\forall \pi^i$,
            \begin{equation}
            V^i(s,(\phi^i, \psi^i, \pi^i)|(\phi^{-i}_*, \psi^{-i}_*, \pi^{-i}_*)) = -1 = V^i(s,(\phi^i_*, \psi^i_*, \pi^i_*)|(\phi^{-i}_*, \psi^{-i}_*, \pi^{-i}_*)).
            \end{equation}
            \item $\forall \psi^i\neq \psi^i_*$ s.t. $\psi^i(s, \mathbf{m}=\{C,C\})=0$,
            \begin{enumerate}
                \item if $\pi^i(s)=C$,
                \begin{equation}
                    V^i(s,(\phi^i, \psi^i, \pi^i)|(\phi^{-i}_*, \psi^{-i}_*, \pi^{-i}_*)) = -3 < -1 = V^i(s,(\phi^i_*, \psi^i_*, \pi^i_*)|(\phi^{-i}_*, \psi^{-i}_*, \pi^{-i}_*)),
                \end{equation}
                \item otherwise $\pi^i(s)=D$,
                \begin{equation}
                    V^i(s,(\phi^i, \psi^i, \pi^i)|(\phi^{-i}_*, \psi^{-i}_*, \pi^{-i}_*)) = -2 < -1 = V^i(s,(\phi^i_*, \psi^i_*, \pi^i_*)|(\phi^{-i}_*, \psi^{-i}_*, \pi^{-i}_*)).
                \end{equation}
            \end{enumerate}
        \end{enumerate}

        \item $\phi^i =\phi^i_*, \psi^i=\psi^i_*, \forall \pi^i$:
        \begin{equation}
             V^i(s,(\phi^i, \psi^i, \pi^i)|(\phi^{-i}_*, \psi^{-i}_*, \pi^{-i}_*)) = -1 = V^i(s,(\phi^i_*, \psi^i_*, \pi^i_*)|(\phi^{-i}_*, \psi^{-i}_*, \pi^{-i}_*)).
        \end{equation}
    \end{enumerate}
    Thus, 
    \begin{equation}
        V^i(s,(\phi^i, \psi^i, \pi^i)|(\phi^{-i}_*, \psi^{-i}_*, \pi^{-i}_*)) \leq V^i(s,(\phi^i_*, \psi^i_*, \pi^i_*)|(\phi^{-i}_*, \psi^{-i}_*, \pi^{-i}_*)),~~~\forall \phi^i, \psi^i, \pi^i.
    \end{equation}
    Therefore, $((\phi^i_*, \psi^i_*, \pi^i_*),(\phi^{-i}_*, \psi^{-i}_*, \pi^{-i}_*))$ is a pure strategy Nash equilibrium in the MCG of Prisoner's Dilemma.
    Meanwhile, $((\phi^i_*, \psi^i_*, \pi^i_*),(\phi^{-i}_*, \psi^{-i}_*, \pi^{-i}_*))$ is also a Pareto-optimal equilibrium. Given the payoff matrix in Table~\ref{PD_matrix}, other possible outcomes are $(-2,-2)$, $(0,-3)$ and $(-3,0)$. Therefore, no further improvement can be made to one player's outcome without reducing the payoff of another player compared to $(-1,-1)$ achieved by $((\phi^i_*, \psi^i_*, \pi^i_*),(\phi^{-i}_*, \psi^{-i}_*, \pi^{-i}_*))$.
\end{proof}
\begin{algorithm}
\caption{Differentiable Commitment Learning (Centralized Version)}\label{alg:dcl_centralized}
\begin{algorithmic}
\State Input: initial action policy parameters $\theta^i$, initial commitment policy parameters $\zeta^i$, initial proposal policy parameters $\eta^i$, initial action-value function parameters $w^i$ for all $i\in \mathcal{N}$.
\For{k=$0,1,2, ...$}
    \State Collect set of trajectories $\mathcal{D}_k=\{\tau_t\}$ by running latest policies $(\theta^i,\zeta^i,\eta^i)$, $\forall i \in \mathcal{N}$.
    \State Compute Monte-Carlo discounted accumulative rewards $\hat{G}_t^i, \forall i\in\mathcal{N}$.
    \State Fit value function for all $i\in\mathcal{N}$ with gradient descent by minimizing the mean-squared error:
    $$w^i_{k+1}=\arg \min_{w^i} \frac{1}{|\mathcal{D}_k|T}\sum_{\tau\in\mathcal{D}_k}\sum_{t=0}^T(Q_{w^i}^i(s_t,\mathbf{a}_t)-\hat{G}_t^i)^2.$$
    \State Estimate action policy gradient for all $i\in\mathcal{N}$ as $$\hat{g}_{\theta^i_k}=\frac{1}{|\mathcal{D}_k|}\sum_{\tau\in\mathcal{D}_k}\sum_{t=0}^T\Big(1-\mathds{1}(\mathbf{c}_t=\mathbf{1})\Big)Q_{w^i_{k+1}}^i(s_t,\mathbf{a}_t)\nabla_{\theta^i_k}\log \pi_{\theta^i_k}^i(a_t^i|s_t).$$
    \State Estimate commitment policy gradient for all $i\in\mathcal{N}$ as 
    \begin{equation*}
    \begin{aligned}
        \hat{g}_{\zeta^i_k}=&\frac{1}{|\mathcal{D}_k|}\sum_{\tau\in\mathcal{D}_k}\sum_{t=0}^T 
        \Big[\mathds{1}(\mathbf{c}_t=\mathbf{1})Q_{w^i_{k+1}}^i(s_t,\mathbf{m}_t)+\Big(1-\mathds{1}(\mathbf{c}_t=\mathbf{1})\Big)Q_{w^i_{k+1}}^i(s_t,\mathbf{a}_t)\Big]  \\
        &\cdot\nabla_{\zeta^i_k}\log\psi^i_{\zeta^i_k}(c^i_t|s_t, \mathbf{m}_t) +\Big[Q_{w^i_{k+1}}^i(s_t,\mathbf{m}_t)-Q_{w^i_{k+1}}^i(s_t,\mathbf{a}_t) \Big]\prod_{j \neq i}\mathds{1}(c^{j}_t=1)\nabla_{\zeta^i}\mathds{1}(c^i_t=1).
    \end{aligned}
    \end{equation*}
    \State Estimate proposal policy gradient w.r.t. the expected return for all $i\in\mathcal{N}$ by
    \begin{equation*}
    \begin{aligned}
        \hat{g}_{\eta^i_k}=&\frac{1}{|\mathcal{D}_k|}\sum_{\tau\in\mathcal{D}_k}\sum_{t=0}^T 
        \Big[\mathds{1}(\mathbf{c}_t=\mathbf{1})Q_{w^i_{k+1}}^i(s_t,\mathbf{m}_t)+\Big(1-\mathds{1}(\mathbf{c}_t=\mathbf{1})\Big)Q_{w^i_{k+1}}^i(s_t,\mathbf{a}_t) \Big]  \\
        &\cdot\Big(\nabla_{\eta^i_k}\log\phi^i_{\eta^i_k}(m^i_t|s_t)+
        \sum_j\nabla_{\eta^i_k}\log\psi^j_{\zeta^j_k}(c^j_t|s_t, \mathbf{m}_t) \Big)\\
        &+\sum_{j}\prod_{l\neq j} \mathds{1}(c^l_t=1)\Big[Q_{w^i_{k+1}}^i(s_t,\mathbf{m}_t)-Q_{w^i_{k+1}}^i(s_t,\mathbf{a}_t) \Big]\nabla_{\eta^i_k}\mathds{1}(c^j_t=1).
    \end{aligned}
    \end{equation*}
    \State Estimate proposal policy gradient w.r.t. incentive-compatible constraints for all $i\in\mathcal{N}$ by
    $$\hat{g}_{\eta^i_k}^{'}=\frac{1}{|\mathcal{D}_k|}\sum_{\tau\in\mathcal{D}_k}\sum_{t=0}^T\sum_j \nabla_{\eta^i_k}\min \{0, Q^j_{w^j_{k+1}}(s_t, \mathbf{m}_t) - Q^j_{w^j_{k+1}}(s_t, \mathbf{a}_t) \}.$$
    \State Update policy parameters for all $i \in \mathcal{N}$ with gradient ascent,
    $$\theta^i_{k+1}=\theta^i_{k}+\beta \hat{g}_{\theta^i_k},$$
    $$\zeta^i_{k+1}=\zeta^i_{k}+\beta \hat{g}_{\zeta^i_k},$$
    $$\eta^i_{k+1}=\eta^i_{k}+\beta \hat{g}_{\eta^i_k} + \lambda\hat{g}_{\eta^i_k}^{'}.$$ 
\EndFor
\end{algorithmic}
\end{algorithm}
\begin{algorithm}
\caption{Differentiable Commitment Learning (Decentralized Version)}\label{alg:dcl_decentralized}
\begin{algorithmic}
\State Input: initial action policy parameters: $\theta^i$, initial estimated action policy parameters of $b \in \mathcal{N}\setminus i$: $\Tilde{\theta}^{ib}$, initial commitment policy parameters: $\zeta^i$, initial estimated commitment policy parameters of $b \in \mathcal{N}\setminus i$: $\Tilde{\zeta}^{ib}$, initial proposal policy parameters: $\eta^i$, initial estimated proposal policy parameters of $b \in \mathcal{N}\setminus i$: $\Tilde{\eta}^{ib}$, initial action-value function parameters: $w^i$ for $i\in \mathcal{N}$, initial estimated action-value function parameters of $b \in \mathcal{N}\setminus i$: $\Tilde{w}^{ib}$ for all $i\in \mathcal{N}$.
    \For{k=$0,1,2, ...$}
    \State Collect set of trajectories $\mathcal{D}_k=\{\tau_t\}$ by running latest policies $(\theta^i,\zeta^i,\eta^i)$, $\forall i \in \mathcal{N}$.
    \State Compute Monte-Carlo discounted accumulative rewards $\hat{G}_t^i, \forall i\in\mathcal{N}$.
    \State Fit value function for all $i\in\mathcal{N}$ with gradient descent by minimizing the mean-squared error:
    $$w^i_{k+1}=\arg \min_{w^i} \frac{1}{|\mathcal{D}_k|T}\sum_{\tau\in\mathcal{D}_k}\sum_{t=0}^T(Q_{w^i}^i(s_t,\mathbf{a}_t)-\hat{G}_t^i)^2.$$
    \State Fit estimated value function of $b$ for $\forall b\in \mathcal{N}\setminus i$ and $\forall i\in\mathcal{N}$ with gradient descent by minimizing the mean-squared error:
    $$\Tilde{w}^{ib}_{k+1}=\arg \min_{w^b} \frac{1}{|\mathcal{D}_k|T}\sum_{\tau\in\mathcal{D}_k}\sum_{t=0}^T(Q_{w^b}^b(s_t,\mathbf{a}_t)-\hat{G}_t^b)^2.$$
    \State Estimate action policy gradient for all $i\in\mathcal{N}$ as 
    $$\hat{g}_{\theta^i_k}=\frac{1}{|\mathcal{D}_k|}\sum_{\tau\in\mathcal{D}_k}\sum_{t=0}^T\Big(1-\mathds{1}(\mathbf{c}_t=\mathbf{1})\Big)Q_{w^i_{k+1}}^i(s_t,\mathbf{a}_t)\nabla_{\theta^i_k}\log \pi_{\theta^i_k}^i(a_t^i|s_t).$$
    \State Estimate action policy of $b$ for $\forall b\in \mathcal{N}\setminus i$ and $\forall i\in\mathcal{N}$ by
    $$\hat{g}_{\Tilde{\theta}^{ib}_k}=\frac{1}{|\mathcal{D}_k|}\sum_{\tau\in\mathcal{D}_k}\sum_{t=0}^T\Big(1-\mathds{1}(\mathbf{c}_t=\mathbf{1})\Big)Q_{\Tilde{w}^{ib}_{k+1}}^b(s_t,\mathbf{a}_t)\nabla_{\Tilde{\theta}^{ib}_k}\log \pi_{\Tilde{\theta}^{ib}_k}^b(a_t^b|s_t).$$
    \State Estimate commitment policy gradient for all $i\in\mathcal{N}$ as 
    \begin{equation*}
    \begin{aligned}
        \hat{g}_{\zeta^i_k}=&\frac{1}{|\mathcal{D}_k|}\sum_{\tau\in\mathcal{D}_k}\sum_{t=0}^T 
        \Big[\mathds{1}(\mathbf{c}_t=\mathbf{1})Q_{w^i_{k+1}}^i(s_t,\mathbf{m}_t)+\Big(1-\mathds{1}(\mathbf{c}_t=\mathbf{1})\Big)Q_{w^i_{k+1}}^i(s_t,\mathbf{a}_t)\Big]  \\
        &\cdot\nabla_{\zeta^i_k}\log\psi^i_{\zeta^i_k}(c^i_t|s_t, \mathbf{m}_t) +\Big[Q_{w^i_{k+1}}^i(s_t,\mathbf{m}_t)-Q_{w^i_{k+1}}^i(s_t,\mathbf{a}_t) \Big]\prod_{j \neq i}\mathds{1}(c^{j}_t=1)\nabla_{\zeta^i}\mathds{1}(c^i_t=1).
    \end{aligned}
    \end{equation*}
    \State Estimate commitment policy gradient of $b$ for $\forall b\in \mathcal{N}\setminus i$ and $\forall i\in\mathcal{N}$ by
    \begin{equation*}
    \begin{aligned}
        \hat{g}_{\Tilde{\zeta}^{ib}_k}=&\frac{1}{|\mathcal{D}_k|}\sum_{\tau\in\mathcal{D}_k}\sum_{t=0}^T 
        \Big[\mathds{1}(\mathbf{c}_t=\mathbf{1})Q_{\Tilde{w}^{ib}_{k+1}}^b(s_t,\mathbf{m}_t)+\Big(1-\mathds{1}(\mathbf{c}_t=\mathbf{1})\Big)Q_{\Tilde{w}^{ib}_{k+1}}^b(s_t,\mathbf{a}_t)\Big] \\
        &\cdot\nabla_{\Tilde{\zeta}^{ib}_k}\log\psi^i_{\Tilde{\zeta}^{ib}_k}(c^b_t|s_t, \mathbf{m}_t) +\Big[Q_{\Tilde{w}^{ib}_{k+1}}^b(s_t,\mathbf{m}_t)-Q_{\Tilde{w}^{ib}_{k+1}}^b(s_t,\mathbf{a}_t) \Big]\prod_{j \neq b}\mathds{1}(c^{j}_t=1)\nabla_{\Tilde{\zeta}^{ib}_k}\mathds{1}(c^b_t=1).
    \end{aligned}
    \end{equation*}
    \State Estimate proposal policy gradient w.r.t. the expected return for all $i\in\mathcal{N}$ by
    \begin{equation*}
    \begin{aligned}
        \hat{g}_{\eta^i_k}=&\frac{1}{|\mathcal{D}_k|}\sum_{\tau\in\mathcal{D}_k}\sum_{t=0}^T 
        \Big[\mathds{1}(\mathbf{c}_t=\mathbf{1})Q_{w^i_{k+1}}^i(s_t,\mathbf{m}_t)+\Big(1-\mathds{1}(\mathbf{c}_t=\mathbf{1})\Big)Q_{w^i_{k+1}}^i(s_t,\mathbf{a}_t) \Big]  \cdot\Big(\nabla_{\eta^i_k}\log\phi^i_{\eta^i_k}(m^i_t|s_t)\\
        &+
        \sum_j\nabla_{\eta^i_k}\log\psi^j_{\zeta^j_k}(c^j_t|s_t, \mathbf{m}_t) \Big)+\sum_{j}\prod_{l\neq j} \mathds{1}(c^l_t=1)\Big[Q_{w^i_{k+1}}^i(s_t,\mathbf{m}_t)-Q_{w^i_{k+1}}^i(s_t,\mathbf{a}_t) \Big]\nabla_{\eta^i_k}\mathds{1}(c^j_t=1).
    \end{aligned}
    \end{equation*}
    \State Estimate proposal policy gradient w.r.t. incentive-compatible constraints for all $i\in\mathcal{N}$ by
\algstore{dcl_decentralized}
\end{algorithmic}
\end{algorithm}
\begin{algorithm}
\begin{algorithmic}
\algrestore{dcl_decentralized}
    \State
    $$\hat{g}_{\eta^i_k}^{'}=\frac{1}{|\mathcal{D}_k|}\sum_{\tau\in\mathcal{D}_k}\sum_{t=0}^T\nabla_{\eta^i_k}\min \{0, Q^i_{w^i_{k+1}}(s_t, \mathbf{m}_t) - Q^i_{w^i_{k+1}}(s, \mathbf{a}) \}+\sum_{b\neq i}\nabla_{\eta^i_k} \min \{0, Q^b_{\Tilde{w}^{ib}_{k}}(s, \mathbf{m}) - Q^b_{\Tilde{w}^{ib}_{k}}(s, \mathbf{a}) \}.$$
    \State Estimate proposal policy gradient w.r.t. the expected return of $b$ for $\forall b\in \mathcal{N}\setminus i$ and $\forall i\in\mathcal{N}$ by
    \begin{equation*}
    \begin{aligned}
        \hat{g}_{\Tilde{\eta}^{ib}_k}=&\frac{1}{|\mathcal{D}_k|}\sum_{\tau\in\mathcal{D}_k}\sum_{t=0}^T 
        \Big[\mathds{1}(\mathbf{c}_t=\mathbf{1})Q_{\Tilde{w}^{ib}_{k+1}}^b(s_t,\mathbf{m}_t)+\Big(1-\mathds{1}(\mathbf{c}_t=\mathbf{1})\Big)Q_{\Tilde{w}^{ib}_{k+1}}^b(s_t,\mathbf{a}_t) \Big]\cdot\Big(\nabla_{\Tilde{\eta}^{ib}_k}\log\phi^b_{\Tilde{\eta}^{ib}_k}(m^b_t|s_t)\\
        &+
        \sum_j\nabla_{\Tilde{\eta}^{ib}_k}\log\psi^j_{\Tilde{\zeta}^{ij}_k}(c^j_t|s_t, \mathbf{m}_t) \Big)+\sum_{j}\prod_{l\neq j} \mathds{1}(c^l_t=1)\Big[Q_{\Tilde{w}^{ib}_{k+1}}^b(s_t,\mathbf{m}_t)-Q_{\Tilde{w}^{ib}_{k+1}}^b(s_t,\mathbf{a}_t) \Big]\nabla_{\Tilde{\eta}^{ib}_k}\mathds{1}(c^j_t=1).
    \end{aligned}
    \end{equation*}
    \State Estimate proposal policy gradient w.r.t. incentive-compatible constraints of $b$ for $\forall b\in \mathcal{N}\setminus i$ and $\forall i\in\mathcal{N}$ by
    $$\hat{g}_{\Tilde{\eta}^{ib}_k}^{'}=\frac{1}{|\mathcal{D}_k|}\sum_{\tau\in\mathcal{D}_k}\sum_{t=0}^T\nabla_{\Tilde{\eta}^{ib}_k}\min \{0, Q^i_{w^i_{k+1}}(s_t, \mathbf{m}_t) - Q^i_{w^i_{k+1}}(s, \mathbf{a}) \}+\sum_{b\neq i} \nabla_{\Tilde{\eta}^{ib}_k}\min \{0, Q^b_{\Tilde{w}^{ib}_{k+1}}(s, \mathbf{m}) - Q^b_{\Tilde{w}^{ib}_{k+1}}(s, \mathbf{a}) \}.$$
    \State 
    Update policy parameters for all $i \in \mathcal{N}$ with gradient ascent,
    $$\theta^i_{k+1}=\theta^i_{k}+\beta \hat{g}_{\theta^i_k},\zeta^i_{k+1}=\zeta^i_{k}+\beta \hat{g}_{\zeta^i_k},\eta^i_{k+1}=\eta^i_{k}+\beta \hat{g}_{\eta^i_k} + \lambda\hat{g}_{\eta^i_k}^{'}.$$ 
    \State Update policy parameters for all $b \in \mathcal{N} \setminus i$ and $i \in \mathcal{N}$ with gradient ascent,
    $$\Tilde{\theta}^{ib}_{k+1}=\Tilde{\theta}^{ib}_k+\beta \hat{g}_{\Tilde{\theta}^{ib}_k},\Tilde{\zeta}^{ib}_{k+1}=\Tilde{\zeta}^{ib}_k+\beta \hat{g}_{\Tilde{\zeta}^{ib}_k},\Tilde{\eta}^{ib}_{k+1}=\Tilde{\eta}^{ib}_k+\beta \hat{g}_{\Tilde{\eta}^{ib}_k} + \lambda\hat{g}_{\Tilde{\eta}^{ib}_k}^{'}.$$
\EndFor
\end{algorithmic}
\end{algorithm}
\section{Hyperparameters}
\label{hyperparam}
For all algorithms, we utilized $2$-layer MLP networks with ReLU activation in the hidden layers. All policy networks apply a softmax function as the output activation, whereas the value network uses a linear output without any activation function. Other hyperparameters are reported in Table~\ref{hyper_pd}, \ref{hyper_grid} and \ref{hyper_ipc}.
\begin{table}[htbp]
\caption{Comparison with Related Frameworks}
\label{table:games}
\centering
\resizebox{1\textwidth}{!}{%
\begin{tabular}{cccccc}
\hline
\textbf{Name}                & \textbf{Commitment} & \textbf{Share Policies} & \textbf{Altruistic Third Party} & \textbf{Reward Transfer} & \textbf{Proposal of Actions} \\ \hline
Commitment Games~\citep{renou2009commitment}             & Unconditional          & Yes                     & No                              & No                       & No                          \\
Conditional Commitment Games~\citep{bryan2010commitment} & Conditional        & Yes                     & No                              & No                       & No                          \\
Contract Mechanism~\citep{hughes2020learning}          & Conditional        & No                      & No                              & No                       & Joint Action                          \\
Formal Contracting~\citep{haupt2022formal}           & Conditional        & No                      & No                              & Yes                      & No                          \\
Mediated-MARL~\citep{ivanov2023mediated}                & N/A                 & No                      & Yes                             & No                       & No                          \\
MCGs (Ours)                  & Conditional        & No                      & No                              & No                       & Self Action                         \\ \hline
\end{tabular}%
}
\end{table}
\begin{table}[htbp]
\caption{Comparison with MARL Baselines}
\label{table:baselines}
\centering
\resizebox{1\textwidth}{!}{%
\begin{tabular}{ccccc}
\hline
\textbf{Name}           & \textbf{Objective} & \textbf{Reward Transfer} & \textbf{Independent Learning} & \textbf{Decentralized Learning} \\ \hline
IPPO~\citep{schulman2017proximal}                            & Individual Returns                     & No              & Yes                  & Yes                    \\
MOCA~\citep{haupt2022formal}                              & Individual Returns                     & Yes             & Yes                  & Yes                    \\
Mediated-MARL~\citep{ivanov2023mediated}                  & Social Welfare + Individual Returns                    & No              & Yes                  & Yes                    \\
Centralized DCL (Ours)                            & Individual Returns                     & No              & No                   & No                    \\
Decentralized DCL (Ours)                           & Individual Returns                     & No              & No                   & Yes                    \\ \hline
\end{tabular}%
}
\end{table}
\begin{table}[ht]
\caption{Hyperparameters of Prisoner's Dilemma} \label{hyper_pd}
\begin{center}
\begin{tabular}{llllll}
\hline
\textbf{Hyperparameters}            & \textbf{DCL} & \textbf{Mediated-MARL} & \textbf{IPPO} & \textbf{MOCA} \\ \hline
Num of Iterations                   & 10,000       & 10,000                 & 10,000        & 10,000        \\
Batch size                          & 128          & 128                    & 128           & 128           \\
Entropy Coef. Start                 & 1.0          & 1.0                    & N/A           & N/A           \\
Entropy Decay                       & 0.0005       & 0.0005                 & N/A           & N/A           \\
Min. Entropy Coef.                  & 0        & 0                  & N/A           & N/A           \\
LR of Value Function                & 8e-4         & 8e-4                   & 8e-4          & 8e-4          \\
LR of Policies                      & 4e-4         & 4e-4                   & 4e-4          & 4e-4          \\
Hidden Layer size                   & 8            & 8                      & 8             & 8             \\
Num of Layers                       & 2            & 2                      & 2             & 2             \\
KL-coefficient                      & N/A          & N/A                    & 0.2           & 0.2           \\
KL-target                           & N/A          & N/A                    & 0.01          & 0.01          \\
Clip Parameter in PPO                      & N/A          & N/A                    & 0.3           & 0.3           \\
Temperature & 10.0         & N/A                    & N/A           & N/A           \\
Temperature Decay                   & 0.05         & N/A                    & N/A           & N/A           \\
Min. Temperature                    & 1.0          & N/A                    & N/A           & N/A \\
Num of Update Per Iteration & 1            & 1                      & 1             & 1             \\
\hline
\end{tabular}
\end{center}
\end{table}
\begin{table}[H]
\caption{Hyperparameters of Grid Game} \label{hyper_grid}
\begin{center}
\begin{tabular}{llllll}
\hline
\textbf{Hyperparameters}            & \textbf{DCL} & \textbf{Mediated-MARL} & \textbf{IPPO} & \textbf{MOCA} \\ \hline
Horizon                             & 16            & 16                    & 16            & 16            \\
Grid Size                           & 4             & 4                     & 4             & 4             \\
Num of Iterations                   & 10,000       & 10,000                 & 10,000        & 10,000        \\
Discount Factor & 0.99       & 0.99                 & 0.99        & 0.99        \\
Batch size                          & 512          & 512                    & 512           & 512           \\
Entropy Coef. Start                 & 2.0          & 2.0                    & N/A           & N/A           \\
Entropy Decay                       & 0.0005       & 0.0005                 & N/A           & N/A           \\
Min. Entropy Coef.                  & 0.001        & 0.001                  & N/A           & N/A           \\
LR of Value Function                & 8e-4         & 8e-4                   & 8e-4          & 8e-4          \\
LR of Policies                      & 4e-4         & 4e-4                   & 4e-4          & 4e-4          \\
Hidden Layer size                   & 32            & 32                      & 32             & 32             \\
Num of Layers                       & 2            & 2                      & 2             & 2             \\
KL-coefficient                      & N/A          & N/A                    & 0.2           & 0.2           \\
KL-target                           & N/A          & N/A                    & 0.01          & 0.01          \\
Clip Parameter in PPO                      & N/A          & N/A                    & 0.3           & 0.3           \\
Temperature & 1.0         & N/A                    & N/A           & N/A           \\
Temperature Decay                   & 0         & N/A                    & N/A           & N/A           \\ 
Min. Temperature                    & 1.0          & N/A                    & N/A           & N/A \\
Num of Update Per Iteration & 30            & 30                      & 30             & 30             \\
\hline
\end{tabular}
\end{center}
\end{table}
\begin{table}[H]
\caption{Hyperparameters of Repeated Pure Conflicting Game} \label{hyper_ipc}
\begin{center}
\begin{tabular}{llllll}
\hline
\textbf{Hyperparameters}            & \textbf{DCL} & \textbf{Mediated-MARL} & \textbf{IPPO} & \textbf{MOCA} \\ \hline
Num of Iterations                   & 10,000       & 10,000                 & 10,000        & 10,000        \\
Discount Factor & 0.99       & 0.99                 & 0.99        & 0.99        \\
Batch size                          & 512          & 512                    & 512           & 512           \\
Entropy Coef. Start                 & 2.0          & 2.0                    & N/A           & N/A           \\
Entropy Decay                       & 0.0005       & 0.0005                 & N/A           & N/A           \\
Min. Entropy Coef.                  & 0.001        & 0.001                  & N/A           & N/A           \\
LR of Value Function                & 8e-4         & 8e-4                   & 8e-4          & 8e-4          \\
LR of Policies                      & 4e-4         & 4e-4                   & 4e-4          & 4e-4          \\
Hidden Layer size                   & 32            & 32                      & 32             & 32             \\
Num of Layers                       & 2            & 2                      & 2             & 2             \\
KL-coefficient                      & N/A          & N/A                    & 0.2           & 0.2           \\
KL-target                           & N/A          & N/A                    & 0.01          & 0.01          \\
Clip Parameter in PPO                      & N/A          & N/A                    & 0.3           & 0.3           \\
Temperature & 1.0         & N/A                    & N/A           & N/A           \\
Temperature Decay                   & 0         & N/A                    & N/A           & N/A           \\ 
Min. Temperature                    & 1.0          & N/A                    & N/A           & N/A           \\ 
Num of Update Per Iteration & 30            & 30                      & 30             & 30             \\
\hline
\end{tabular}
\end{center}
\end{table}
\section{Many-player Experiments}
\label{additional_exps}
To investigate how DCL handles scalability with many players, we conducted experiments on an $N$-player public goods game. For each agent $i$, the reward is calculated as $R^i = \sum_j C^j * \beta - C^i$, where $C^i$ denotes the contribution of agent $i$, $\beta$ denotes the benefit factor with a range between $(1,N)$. In our experiments, we set $\beta=1.5$ for all scenarios. 
Results in Table~\ref{NPD} indicate that DCL  with incentive-compatible constraints scales effectively with large numbers of agents. While the runtime of DCL increases with the number of agents, the agreement rate of joint proposals remains stable ($>0.99$), achieving high social welfare.
\begin{table}[H]
\caption{DCL-IC on Many-player Public Goods Game} \label{NPD}
\centering
\begin{tabular}{cccc}
\hline
Number of Agents & Run Time (Hours) & Agreement Rate                & Social Welfare                 \\ \hline
2                & 4                & $0.996\pm 0.002$ & $0.997 \pm 0.002$ \\
3                & 7                & $0.994\pm0.001$  & $1.491\pm0.004$   \\
5                & 12               & $0.996\pm0.001$  & $1.989\pm 0.002$  \\
10               & 32               & $0.991\pm0.001$  & $3.659\pm 0.143$  \\ \hline
\end{tabular}
\end{table}
\section{Other Related Works}
\subsection{Cooperation Problems in Mixed-Motive Environments}
The causes of cooperation failures between self-interested agents in mixed-motive environments have been primarily categorized into two types: information problems and commitment problems~\citep{dafoe2020open, powell2006war, fearon1995rationalist}. 
Information problems refer to cooperation failures caused by incorrect or insufficient information, which frequently occur in partially observable environments. Existing works have demonstrated that information problems can be alleviated by communication~\citep{kim2020communication, sukhbaatar2016learning, foerster2016learning} and opponent reasoning~\citep{konan2022iterated,jaques2019social, wen2019probabilistic}.
However, in mixed-motive environments, agents driven by conflicting self-interests may deceive others regarding their private observations \citep{lin2024information,kamenica2019bayesian,ID2019, IDsurvey}. 
Cooperation may also fail due to agents' inability to make credible commitments, known as commitment problems, even in the absence of information asymmetries. For instance, cooperation can not be achieved through cheap talk communication or non-binding promises of cooperation in the Prisoner’s Dilemma~\citep{rapoport1965prisoner}, as agents achieve higher payoffs by defecting regardless of the opponent's actions. To address commitment problems, a commitment device is often required to ensure that agents fulfill their commitments, either by restricting their actions or imposing penalties for noncompliance~\citep{sun2023cooperative,165211}. 
Static conditional commitments facilitated by such devices have been shown to enhance cooperation in the prisoner's dilemma~\citep{kalai2010commitment,renou2009commitment,schelling1980strategy}. However, these fixed strategies are difficult to generalize across various games and environments.
\subsection{Comparison with Related Works}
Table~\ref{table:games} and \ref{table:baselines} summarize the differences and similarities between various types of games and associated algorithms to optimize strategies.

\end{document}